\documentclass[journal,12pt,onecolumn]{IEEEtran}

\usepackage{mathptmx}   
\usepackage{amsfonts, amssymb, amsmath}
\usepackage{epsfig}
\usepackage{theorem}
\usepackage{textcomp}
\usepackage{tipa}
\usepackage{multirow}
\usepackage{tabularx}
\usepackage[noadjust]{cite}
\makeatletter
\let\NAT@parse\undefined
\makeatletter
\usepackage{natbib}

\let\chapter\section 
\usepackage[linesnumbered,ruled]{algorithm2e}
\SetKwInput{KwIn}{Initial condition}
\SetKwInput{KwResult}{Outcome}

\newtheorem{theorem}{Theorem}

\newtheorem{proposition}[theorem]{Proposition}

\newtheorem{problem}{Problem}

\newcommand{\qed}{\hfill $\Box$\\}

\def\vec#1{\mathbf{#1}}

\def\otp{{OTP}}
\def\bmt{{BMT}}
\def\rmt{{RMT}}

\title{\LARGE \bf Optimal Tourist Problems and Anytime Planning of Trip Itineraries }
\author{Jingjin Yu \qquad Javed Aslam \qquad Sertac Karaman \qquad Daniela Rus%
\thanks{J. Yu and D. Rus are with the Computer Science and Artificial Intelligence Lab at the Massachusetts Institute of Technology. E-mail: \{jingjin, rus\}@csail.mit.edu. J. Aslam is with the Department of Computer Science at Northeastern University. Email: jaa@ccs.neu.edu. S. Karaman is with the Department of Aerospace and Astronautics Engineering at the Massachusetts Institute of Technology. E-mail: karaman@mit.edu.}%
\thanks{This work was supported in part by ONR projects N00014-12-1-1000 and N00014-09-1-1051, and the Singapore-MIT Alliance on Research and Technology (SMART) Future of Urban Mobility project.}%
}

\begin{document}
\maketitle
\begin{abstract}We introduce and study the problem in which a mobile sensing robot (our tourist) is tasked to travel among and gather intelligence at a set of spatially distributed point-of-interests (POIs). The quality of the information collected at each POI is characterized by some non-decreasing reward function over the time spent at the POI. With limited time budget, the robot must balance between spending time traveling to POIs and spending time at POIs for information collection (sensing) so as to maximize the total reward. Alternatively, the robot may be required to acquire a minimum mount of reward and hopes to do so with the least amount of time. We propose a mixed integer programming (MIP) based anytime algorithm for solving these two NP-hard optimization problems to arbitrary precision. The effectiveness of our algorithm is demonstrated using an extensive set of computational experiments including the planning of a realistic itinerary for a first-time tourist in Istanbul. 
\end{abstract}

\section{Introduction}\label{section:introduction}
Imagine that a roboticist travels to Turkey to attend an international conference in Istanbul. Unfortunately, due to her busy schedule, our roboticist does not have much time for touring this historic city. Yet, as luck would have it, near the end of her trip, she finds herself with a day of spare time and decides to do some sightseeing. Planning such a day trip, however, turns out to be quite challenging: the roboticist must decide among a large number of point-of-interest (POIs) which ones to go to, how to travel from one POI to another, and how much time she should spend at each POI that she does decide to visit. Naturally, she hopes to get the most out of her tour under her limited time budget. Could we help our roboticist plan an optimal itinerary for such a journey automatically? 

Alternatively, an environmental scientist may need to plan an automated, GPS-guided trip for an aerial mobile (sensing) robot to collect scientific data at a set of spatially distributed locations. Because of the high cost associated with operating the robot, our scientist, similar to our roboticist in Istanbul, must select a subset of locations for the aerial robot to visit and decide how much effort (time) the robot should spend at each location to perform necessary measurements. Is there a principled method that our environmental scientist can use for planning such a trip with optimality guarantees? 

In this paper, we propose the Optimal Tourist Problem (\otp) that is motivated by and models after the scenarios mentioned above. In the basic setup, a tourist is interested in visiting some $n$ POIs that are spatially distributed. Each POI is associated with a {\em reward function} or {\em learning curve} that is non-decreasing over the time spent at the POI. Because traveling between POIs and staying at a POI to gain reward are both time consuming, optimization problems naturally arise. We introduce two such related problems. In the first problem, a {\em reward-maximizing tourist} (\rmt) seeks to maximize the gained reward given limited time budget. From a dual perspective, in the second problem, a {\em budget-minimizing tourist} (\bmt) seeks to minimize the time spent to collect a predetermined amount of reward. We provide a mixed integer programming (MIP) based {\em anytime} algorithm for solving both \rmt\, and \bmt\, variants of the \otp\, problem. 

The primary motivation behind our study of \otp\, is its potential application to robotic surveillance and monitoring problems such as automated reconnaissance and scientific survey \cite{SmiSchSmiJonRusSuk11,GroKelKumPap06}, which we refer to under the umbrella term of {\em informative path planning} (IPP). In an IPP problem, a path is planned to satisfy some information collection objective, sometimes under additional constraints such as path length or total time limit. In \cite{AlaFatSmi14}, an $O(\log n)$ approximation algorithm yields iterative TSP paths that minimize the maximum latency (the inverse of the frequency with with a node is visited) across all $n$ nodes in a connected network. In \cite{SmiSchRus12}, the authors proposed a method for generating speed profiles along predetermined cyclic (closed) paths to keep bounded the uncertainty of a varying field using single or multiple robots. For the problem of observing stochastically arriving events at multiple locations with a single mobile robot, a $(1+\epsilon)$-optimal algorithm was proposed in \cite{YuKarRus14ICRA} to solve the multi-objective optimization problem of maximizing event observation in a balanced manner and minimizing delay between event observations across the locations. Recently, a method called {\em Recursive Adaptive Identification} is proposed as a polynomial time polylogarithmic-approximation algorithm for attacking adaptive IPP problems \cite{LimHsuLee14}. Sampling based methods \cite{KavSveLatOve96,Lav98c,KarFra11IJRR} have also been applied to IPP problems with success.  In \cite{HolSuk13}, Rapidly-Exploring Random Graphs (RRG) are combined with branch-and-bound methods for planning most informative paths. In \cite{LanSch13}, the authors tackle the problem of planning cyclic trajectories for the best estimation of a time-varying Gaussian Random Field, using a variation of RRT called Rapidly-Expanding Random Cycles (RRC). 

An optimization problem that is intimately connected to \otp\, is the Orienteering Problem (OP) \cite{ChaGolWas96a, VanSouVan11, GavKonMasPan14}, which is obtained when rewards at the POIs are fixed in an \rmt\, problem. The fixed reward is collected in full once a POI is visited. OP, which is easy to see as an NP-hard problem, is observed to be difficult to solve exactly for even medium sized instances with over a hundred of POIs. On the side of approximation algorithms, constant approximation ratios down to $(2 + \epsilon)$ are only known under metric settings for OP with uniform reward across the POIs on undirected graphs \cite{CheKorPal12}. No constant ratio approximation algorithm is known for directed graphs. On the other hand, many MIP-based algorithms exist for OP and related problems \cite{VanSouVan11, GavKonMasPan14}. These algorithms often allow the precise encoding of the problem in the MIP model. A work in this domain that is closest to ours studies an OP problem in which the reward may depend on the time spent at the POIs \cite{ErdLap13}. It proposes a solution method that iteratively adds constraints that are violated by the incomplete model. In comparison, our work studies a more general problem that allows multiple starting POIs and arbitrary reward functions. Moreover, we construct a static ({\em i.e.} constraints are fixed), arbitrarily precise  MIP model that gives rise to a natural anytime algorithm. 

On the side of trip planning problems, many interesting works \cite{ChoFelAme10,BasDasAmeYu11,HyoZheXieWoo12} compute ``optimal'' itineraries according to some reward metric. For example, the authors of~\cite{ChoFelAme10} apply a recursive greedy approximation algorithm for OP \cite{ChePal05} to plan suggested itineraries. Most of these work focus on the data mining aspect of trip planning problems, {\em e.g.}, how POI related data, such as the average visiting times for POIs and tourist preference through POI correlations, may be derived and used. In contrast, we provide a clean separation between two elements of the \otp\, problem, the transportation model and the reward model, and focus on the interaction between these two elements through an algorithmic study. 

The rest of the paper is organized as follows. In Section~\ref{section:formulation}, we formulate the two variants of \otp, \rmt\, and \bmt. In Section~\ref{section:mip-model}, we provide a step-by-step introduction of our MIP model for solving the proposed \otp\, variants, after which many generalizations are also presented. In Section~\ref{section:algorithm}, we discuss the overall algorithm and some of its important properties in more detail. We present computational simulations in Section~\ref{section:experiment} and conclude in Section~\ref{section:conclusion}. 

\section{Problem Formulation}\label{section:formulation}
Let the set $V = \{v_1, \ldots, v_n\}$ represents $n$ {\em point-of-interests} (POIs) in $\mathbb R^2$. There is a {\em directed edge} $e_{i,j}$ between two POI vertices $v_i, v_j \in V$ if there is a path from $v_i$ to $v_j$ that does not pass through any intermediate POIs. When an edge $e_{i,j}$ exists, let $d_{i,j}$ denote its length. There is a tourist (alternatively, an agent or a mobile robot) that travels between the POIs following single integrator dynamics. Denoting the tourist's location as ${\vec{p}}$, when the tourist is traveling from POI to POI, $\dot{\vec{p}} = u, \parallel u\parallel = 1$. Otherwise, $\dot{\vec{p}} = 0$. 

The tourist is interested in visiting the POIs. To do so, she starts from some {\em base} vertex $v_B \in B\subset V$ with $\vert B \vert = n_B \le n$, travels between the POIs, and eventually returns to $v_B$. For example, $B$ may represent the choices of hotels. For each $v_i \in V$, she associates a maximum {\em reward} $r_i$ with the location, which can be gained through spending time at $v_i$. We assume that the obtained reward depends on the time $t_i$ the tourist spends at $v_i$. More precisely, the obtained reward is defined as $r_i f_i(t_i)$, in which $f_i \in [0, 1]$ is some function of $t_i$ that is non-decreasing. We further require that $f_i$ is $C^1$ continuous and $f_i'(0)$ is bounded away from zero. That is, for all $1 \le i \le n$, $f_i'$ is continuous and $f_i'(0) \ge \lambda$ for some fixed $\lambda > 0$. We also assume that $f(0) = 0$ for convenience (it can be easily verified later that this does not reduce generality).  

\textbf{Remark.} We mention that no generality is lost by focusing on non-decreasing functions. After presenting our MIP models in Section~\ref{section:mip-model}, it will become clear that any reasonable $f_i$ can be turned into an equivalent non-decreasing function which can then be used in setting up the MIP model. We will revisit this point in Section~\ref{subsection:generalization}.

The function $f_i$ may effectively be viewed as a {\em learning curve}. In this paper, two specific types of one-parameter learning curves are studied in detail: {\em linear} and {\em exponential}. Let $\lambda_i > 0$ denote the {\em learning rate}. In the case of a linear learning curve,  
\begin{align}\label{equation:linear}
f_i(t_i) =\lambda_it_i,\quad 0 \le t_i \le \frac{1}{\lambda_i}.
\end{align}

The exponential learning curve is specified as
\begin{align}\label{equation:exponential}
f_i(t_i) =1 - e^{-\lambda_it_i},\quad 0 \le t_i \le +\infty, 
\end{align}
which captures the notion of ``diminishing return'' that are often present in learning tasks. 

After a trip is completed, our tourist would have traveled through a subset of the edges $E_{tr} \subset E$ and have spent time $t_1, \ldots, t_n, t_i \ge 0$ at the $n$ POIs. She would have spent a total time of 
\begin{align}\label{equation:total-time}
J_T := \sum_{e_{i,j} \in E_{tr}}d_{i,j} + \sum_{i=1}^{n} t_i
\end{align}
and gained a total reward of 
\begin{align}\label{equation:total-reward}
J_R := \sum_{i = 1}^n r_if_i(t_i). 
\end{align}

Note that some edges $e_{i,j}$ may be passed through by the tourist multiple times, in which case $d_{i,j}$ is included once each time $e_{i,j}$ is enumerated in~\eqref{equation:total-time}. That is, $E_{tr}$ is a multi-set. We define $T := \{t_1, \ldots, t_i\}$, $R: = \{r_1, \ldots, r_n\}$, and $F := \{f_1, \ldots, f_n\}$. 

During the trip planning phase, a tourist often faces the challenging task of planning ahead so as to spend the optimal amount of time to travel and to do sightseeing to gain the most out of a trip. This gives rise to two \otp\, variants. In the first, our optimal tourist is given a time budget $M_T$, during which she hopes to maximize her total reward. That is, 

\begin{problem}[Reward-Maximizing Tourist (\rmt)]\label{problem:max-reward} Given a 5-tuple $(V, B, D, R, F)$ and a time budget $M_T > 0$, compute the sets $E_{tr}$ and $T$ such that $J_R$ is maximized under the constraint $J_T \le M_T$. 
\end{problem}

We do not need to specify the edge set $E$ because it is implicitly fixed by $D$. The second, equally natural problem is in a sense a dual problem of \rmt, in which the goal is to minimize the time spent to achieve a predetermined reward.

\begin{problem}[Budget-Minimizing Tourist (\bmt)]\label{problem:min-time} Given a 5-tuple $(V, B, D, R, F)$ and a reward requirement $M_R > 0$, compute the sets $E_{tr}$ and $T$ such that $J_T$ is minimized under the constraint $J_R \ge M_R$. 
\end{problem}

Besides \rmt\, and \bmt\, as formulated in this section, many practical variations are possible. For example, it may be the case that a path (starting and ending at hotels, train stations, and so on) is required instead of a closed tour. Alternatively, maybe a multi-day itinerary is more desirable than a one-day itinerary. These variations and a few additional generalizations are also addressed later in this paper (in Section~\ref{subsection:generalization}). 

\textbf{Remark.} We emphasize that the problems formulated in this section apply to an array of scenarios other than itinerary planning for tourists. For example, our tourist may well be a mobile aerial robot equipped with on-board cameras and automated computer vision-based algorithms for traffic monitoring at key intersections in a large city. In this case, spending more time at a given location will allow more observations, leading to higher quality information about the traffic pattern at the given location. Given limited flying time, the aerial robot must balance between traveling around and spending time at important sites to gather more traffic information (under some proper metric). We can easily imagine extensions of this traffic monitoring application to surveillance, reconnaissance, and scientific exploration tasks. 

\section{MIP Models for \bmt\, and \rmt}\label{section:mip-model}
In this section, we propose mixed integer programming models for solving \rmt\, and \bmt\, using an MIP solver. First, we describe an MIP model derived from an existing one for the orienteering problem (OP) that applies to \rmt\, and \bmt\, problems with $\vert B \vert = 1$ ({\em i.e.}, a single base) and linear learning curves. The case of $\vert B\vert = 1$ is often referred to as a {\em rooted} problem. Then, the MIP model is generalized to allow multiple bases and arbitrary learning curves through linearization. Before moving to model construction, we point out that the proposed problems are computationally intractable, given their similarity to TSP and OP. 

\begin{proposition}\rmt\, and \bmt\, are NP-hard.
\end{proposition}
\noindent{\sc Proof.} Let $r_i \equiv 1$ and let the functions from the set $F$ be linear with unit slope, {\em i.e.}, $f_i' = \lambda_i \equiv 1$. The maximum achievable reward is then $n$ and achieving such a reward requires $t_i = 1$ for all $1 \le i \le n$. Under these restrictions, solving a \bmt\, instance with $M_R = n$ is equivalent to finding a TSP tour over all $n$ POI vertices, which is NP-hard. Now, given a time budget $M_T$, the decision problem of whether $M_T$ is sufficient for achieving a reward of $J_R \ge n$ is NP-hard, implying that \rmt\, is NP-hard as well. ~\qed 

\subsection{MIP Model for a Single Base and Linear Learning Curves}\label{subsection:basic-model}
In this subsection, we introduce an MIP model for \bmt\, and \rmt\, with a single base and with the set $F$ being linear functions. These models are partially based on models from \cite{VanSouVan11,ErdLap13}. Without loss of generality, let our tourist start from $v_1$. Because the reward at a given POI only depends on the total time spent at the POI, we also assume that the time the tourist spent at a POI is spent during a single visit to the POI. When a tourist spends time at a POI, we say the tourist {\em stays} at the POI. With these assumptions, the tourist will eventually have stayed at some $\ell$ POIs with the order $v_{s_1}, \ldots, v_{s_{\ell}}$, and have spent time $t_{s_1}, \ldots, t_{s_{\ell}}$ at these POIs. For $i \notin \{s_1, \ldots, s_{\ell}\}$, $t_i = 0$. 

Although the tourist only needs to stay at a POI at most once, she may need to pass through a POI multiple times ({\em e.g.}, if the POI is a transportation hub). To distinguish these two types of visits to a POI, we perform a transitive closure on the set $D$. That is, we compute all-pairs shortest paths for $v_i, v_j \in V, 1 \le i, j \le n$. This gives us a set of shortest directed paths $P = \{p_{i,j}\}$ with corresponding lengths $D' = \{d_{i,j}'\}$. We say that the tourist {\em takes} a path $p_{i,j}$ if the tourist stays at $v_j$ immediately after staying at $v_i$, except when the tourist starts and ends her trip at $v_1$. With this update, the tourist's final tour is  simply $p_{s_1, s_2}, \ldots, p_{s_{\ell}, s_1}$. Let $x_{ij}$ be a binary variable with $x_{ij} = 1$ if and only if $p_{i,j}$ is taken by the tourist. 

The number of times that the tourist stays at (resp. leaves after staying) a POI vertex $v_i$ is $\sum_{j = 1, j \ne i}^n x_{ij}$ (resp. $\sum_{j = 1, j \ne i}^n x_{ji}$). Both summations can be at most one since by assumption, the tourist never stays at a POI twice. The tour constraint then says they must be equal, {\em i.e.}, $\sum_{j = 1, j \ne i}^n x_{ij} = \sum_{j = 1, j \ne i}^n x_{ji}$. Let $x_i$ be the binary variable indicating whether the tourist stayed at $v_i$. We have the following edge-use constraints
\begin{align}\label{equation:edge-use}
\displaystyle\sum_{j = 1, j \ne i}^n x_{ij} = \sum_{j = 1, j \ne i}^n x_{ji} = x_i\le 1, \quad \forall 2 \le i \le n.
\end{align}
The case of $i = 1$ is special since we need to ensure that $v_1$ is visited, even if the tourist does not actually {\em stay} at $v_1$. For this purpose, we add a self-loop variable $x_{11}$ at $v_1$ and require
\begin{align}\label{equation:edge-use-base}
\displaystyle\sum_{j = 1}^n x_{1j} = \sum_{j = 1}^n x_{j1} = x_1 = 1.
\end{align}
The constraints~\eqref{equation:edge-use} and~\eqref{equation:edge-use-base} guarantee that the tourist takes a tour starting from $v_1$. However, they do not prevent multiple disjoint tours from being created. To prevent this from happening, a {\em sub-tour restriction} constraint is introduced. Let $2 \le u_i \le n$ be integer variables for $2 \le i \le n$. If there is a single tour starting from $v_1$, then $u_i$ can be chosen to satisfy the constraints
\begin{align}\label{equation:no-sub-tour}
\displaystyle u_i - u_j + 1 \le (n - 1)(1 - x_{ij}),\quad 2 \le i, j \le n, i \ne j.
\end{align}

To see that this is true, note that since $u_i - u_j + 1 \le n - 1$ regardless the of the values taken by $2 \le u_i, u_j \le n$, \eqref{equation:no-sub-tour} can only be violated if $x_{ij} = 1$. The condition $x_{ij} = 1$ only holds if the path $p_{i,j}$ taken. Setting $u_i$ to be the order with which the tourist stays at $v_i$, if $x_{ij} = 1$, then $u_i - u_j + 1 = 0$, satisfying~\eqref{equation:no-sub-tour}. On the other hand, if there is another tour besides the one starting from $v_1$ and when $v_{ij} = 1$, then the RHS of~\eqref{equation:no-sub-tour} equals zero. For~\eqref{equation:no-sub-tour} to hold, we must have $u_i - u_j + 1 \le 0 \Rightarrow u_i < u_j$. However, this condition cannot hold for all consecutive pairs of POI vertices on a cycle. Thus,~\eqref{equation:no-sub-tour} enforces that only a single tour may exist. 

With the introduction of the variables $\{x_{ij}\}$, the time spent by the tourist is given by
\begin{align}\label{equation:time-objective}
J_T = \sum_{i = 1}^n\sum_{j = 1, j \ne i}^n x_{ij}d_{ij} + \sum_{i = 1}^n t_i.
\end{align}

To represent the total reward $J_R$, we introduce a continuous variable $w_i, 1 \le i \le n$, to denote the reward collected at $v_i$. For a linear $f_i$, $\lambda_i$, the learning rate, is simply the slope of $f_i$. The reward $w_i$ and the visiting time $t_i$ then satisfy
\begin{align}
w_i \le r_ix_i, \label{equation:reward-max} \\
w_i = t_i\lambda_i, \label{equation:reward-proportion}
\end{align}

The constraint~\eqref{equation:reward-max} allows reward only if the tourist stays at $v_i$ and limits the maximum reward at $r_i$. The constraint~\eqref{equation:reward-proportion} reflects the linear dependency of the reward $w_i$ over the visiting time $t_i$. The total reward $J_R$ is simply 

\begin{align}\label{equation:reward-objective}
J_R = \sum_{i = 1}^n w_i.
\end{align}

Solving \rmt\, with a single base and linear learning curves can then be encoded as a mixed integer program that seeks to maximize $J_R$ subject to $J_T \le M_T$,~\eqref{equation:edge-use},~\eqref{equation:edge-use-base},~\eqref{equation:no-sub-tour},~\eqref{equation:reward-max}, and~\eqref{equation:reward-proportion}. Similarly, solving \bmt\, with a single base and linear learning curves can be encoded as a mixed integer program that minimizes $J_T$ subject to $J_R \ge M_R$,~\eqref{equation:edge-use},~\eqref{equation:edge-use-base},~\eqref{equation:no-sub-tour},~\eqref{equation:reward-max}, and~\eqref{equation:reward-proportion}. 

\subsection{Incorporating Multiple Bases}\label{subsection:multiple-base}

\begin{figure}[htp]
\begin{center}
    \includegraphics[width=2.8in]{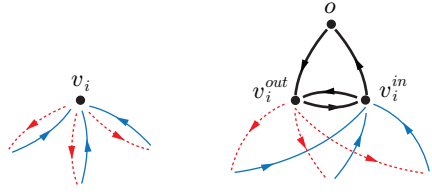} 
\end{center}
\caption{\label{figure:gadget} [left] A base vertex $v_i$ and its outgoing (dotted) and incoming (solid) edges. [right] The gadget that split $v_i$ into $v_i^{in}$ and $v_i^{out}$, along with the split edges and the newly added four (bold) edges.}
\end{figure}

We now look at the case of $\vert B \vert > 1$. To enable the selection of any particular $v_i \in B$, a virtual {\em origin} vertex $o$ is created, which is both a source and a sink. Then, each base vertex $v_i$ is split into two copies, $v_{i}^{in}$ and $v_{i}^{out}$. The edges connecting $v_i$ to other POI vertices of $V$ are split such that all edges going from $v_i$ to other POI vertices are now rooted at $v_i^{out}$ and all edges connecting other POI vertices to $v_i$ are now ending at $v_i^{in}$. In addition, two crossover edges between $v_{i}^{in}$ and $v_{i}^{out}$ are added, one in each direction. Lastly, an outgoing edge from $o$ to $v_i^{out}$ and an incoming edge from $v_i^{in}$ to $o$ are added. An illustration of this gadget is given in Figure~\ref{figure:gadget}.

This gadget is duplicated for every element of $B$ using the same origin vertex $o$. The basic MIP model from the previous subsection is then updated to enable the routing of the tourist through at least one element of $B$. For each $v_i \in B$, we create four additional binary variables to represent whether the four newly added edges are used in a solution. These variables are $x_i^{o,out}$ (edge from $o$ to $v_i^{out}$), $x_{i}^{in,o}$ (edge from $v_i^{in}$ to $o$), $x_i^{out,in}$ (edge from $v_i^{out}$ to $v_i^{in}$), and $x_i^{in,out}$ (edge from $v_i^{in}$ to $v_i^{out}$). To ensure that at least one vertex of $B$ is used, we add the constraint
\begin{align}\label{equation:base}
\sum_{v_i \in B} x_i^{o,out} = 1.
\end{align}

The edge-use constraints also need to be updated accordingly. Due to the vertex split for vertices from the set $B$, we have two sets of such edge-use constraints. The constraint~\eqref{equation:edge-use} now applies to all non-base vertices. The constraint~\eqref{equation:edge-use-base} is updated for all base vertices $v_i \in B$ to
\begin{align}
\sum_{j=1, j\ne i}^n x_{ij} + x_i^{out,in} - x_i^{in, out} - x_i^{o, out} = 0, \label{equation:base-1} \\
\sum_{j=1, j\ne i}^n x_{ji} + x_i^{out,in} - x_i^{in, out} - x_i^{in, o} = 0. \label{equation:base-2}
\end{align}

With constraint~\eqref{equation:base}, $o$ goes to exactly one $v_i^{out}$ and later returns to $v_i^{in}$. Then, constraints~\eqref{equation:base-1} and~\eqref{equation:base-2}, along with the existing edge-use constraint~\eqref{equation:edge-use}, ensure that one or more tours are created. Finally, to prevent multiple tours from being created, we update the variables $u_i$'s to $1 \le u_i \le n$ for $1 \le i \le n$. For a base vertex $v_i \in B$, we add the constraint
\begin{align}\label{equation:sub-tour-nb}
u_i - u_j + 1 \le (2 - x_{ij} - x_i^{in,out})n.
\end{align}

If $v_i$ is not a base vertex, we require
\begin{align}\label{equation:sub-tour-n}
u_i - u_j + 1 \le (1 - x_{ij})n.
\end{align}

Constraints~\eqref{equation:sub-tour-nb} and~\eqref{equation:sub-tour-n} replace the constraint~\eqref{equation:no-sub-tour}. The constraint~\eqref{equation:sub-tour-n} has the same effect as the constraint~\eqref{equation:no-sub-tour} in preventing a separate tour from being created. For base vertices, when $x_i^{in,out} = 1$, which is the case unless $x_i^{o,out} \ne 1$, the constraint~\eqref{equation:sub-tour-nb} is the same as~\eqref{equation:sub-tour-n}. If $x_i^{o,out} = 1$, then~\eqref{equation:sub-tour-nb} becomes $u_i - u_j + 1 \le n + (1-x_{ij})n$, which always holds. That is, the constraint~\eqref{equation:sub-tour-nb} treats the selected base vertex differently. 

\subsection{Linearization of Arbitrary Learning Curves}\label{subsection:linearization}
To accommodate arbitrary learning curves into our MIP model, a linearization scheme is used. We show that, with carefully constructed linear approximations of $f_i$'s, arbitrarily optimal MIP models can be built. 

\begin{figure}[htp]
\begin{center}
    \includegraphics[width=3.15in]{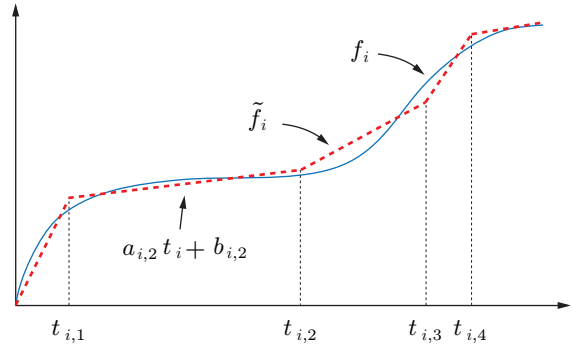} 
\end{center}
\caption{\label{figure:approximation} Approximation of some $f_i$ with a continuous, piecewise linear function (bold dashed line segments). The approximation is concave between $[0, t_{i,2}]$, $[t_{i,2}, t_{i,3}]$, and so on.}
\end{figure}

The basic idea behind our linearization scheme is rather simple. Given a $C^1$ continuous $f_i \in [0, 1]$ with $f_i'(0) \ge \lambda > 0$, it can be approximated to arbitrary precision with a continuous, piecewise linear function $\widetilde{f}_i$ such that for arbitrary $\epsilon > 0$ and all $t_i \ge 0$,
\begin{align}\label{equation:approximation-2}
\frac{\vert f_i - \widetilde{f_i} \vert}{f_i} \le \epsilon,
\end{align}
with $\widetilde{f}_i$ having the form (see, {\em e.g.}, Figure~\ref{figure:approximation})
\begin{align}\label{equation:approximation-1}
\widetilde{f}_i = \left\{
\begin{array}{llc}
a_{i,1}t_i + b_{i,1}, && 0 \le t_i \le t_{i,1} \\
a_{i,2}t_i + b_{i,2}, && t_{i,1} \le t_i \le t_{i,2} \\
\ldots,&& \ldots \\
a_{i,k_i}t_i + b_{i,k_i}, && t_{i,k_i - 1} \le t_i \le \infty
\end{array}
\right.
\end{align}
A numerical procedure for computing such an $\widetilde{f}_i$ is provided in Section~\ref{section:algorithm}.

Once a particular $\widetilde{f_i}$ is constructed, the constraints on the reward $w_i$ must be updated. To make the explanation clear, we use the $\widetilde{f_i}$ from Figure~\ref{figure:approximation} as a concrete example.  Starting from $t_i = 0$, we introduce a new continuous variable $t_i^1$ over the first maximally concave segment of $f_i$. In the case of the $\widetilde{f}_i$ in Figure~\ref{figure:approximation}, the first maximally concave segment contains two line segments, ending at $t_{i,2}$. In this case, we have 
\[0 \le t_i^1 \le t_{i,2}.\] 

To represent the reward obtained over the first maximally concave segment, a continuous variable $w_i^1$ is introduced, which satisfies the following constraints
\[
w_i^1 \le a_{i,1}t_i^1 + b_{i,1},\quad
w_i^1 \le a_{i,2}t_i^1 + b_{i,2}.
\]

Then, for the next maximally concave segment, another continuous variable $t_i^2$ is introduced. In our example, the second maximally concave segment contains one line segment and thus
\begin{align}\label{equation:ti2} 
t_{i,2} \le t_i^2 \le t_{i,3}.
\end{align}

We need to ensure that $t_i^2$ is active only if $t_i^1$ is maximized. We achieve this through the introduction of an additional binary variable $x_{i}^2$, which is set to satisfy the constraint
\[ x_i^2 \le \frac{t_i^1}{t_{i,2}}. \]

The constraint ensures that $x_i^2 = 1$ only if $t_i^1$ is maximized and takes the value $t_{i,2}$. To avoid potential numerical issues that may prevent $x_i^2 = 1$ from happening, in practice, we may write the constraint as $x_i^2 \le (t_i^1 + \delta)/t_{i,2}$, in which $\delta$ is a small positive real number. We can then activate $t_i^2$ through the constraint 
\[ t_i^2 \le x_i^2(t_{i,3} - t_{i,2}) + t_{i,2}, \]
which also renders the constraint~\eqref{equation:ti2} unnecessary. The reward for this second maximally concave segment, $w_i^2$, is then
\[w_i^2 \le a_{i,3}t_i^1 + b_{i,3} - (a_{i,2}t_{i,2} + b_{i,2}). \]

After all of $\widetilde{f}_i$ are encoded as such, we combine the individual time and reward variables into $t_i$ and $w_i$ as
\begin{align}
t_i = t_i^1 + (t_i^2 - t_{i,2}) + \ldots, \\
w_i = w_i^1 + w_i^2 + \ldots. \qquad
\end{align}

We note that the additional constraints that are introduced is proportional to the complexity of $\widetilde{f_i}$. We now prove that the overall MIP model constructed in this way allows arbitrary approximations of the original problem. 

\begin{theorem}Given an \rmt\, instance specified by a 5-tuple $(V, B, D, R, F)$, $M_T > 0$, and a positive real number $\epsilon$, a $(1 + \epsilon)$-optimal solution of this \rmt\, instance can be computed by solving a mixed integer programming problem, obtained over a $(1 + \epsilon/2)$ piece-wise linear approximation of $F$. 
\end{theorem}
\noindent{\sc Proof.} Assume that the \rmt\, instance has an optimal solution that has a reward $J_R^*$ and spends $t_1^*, \ldots, t_n^*$ time at the $n$ POI vertices. Let $\widetilde{f}_i$ be a piece-wise linear $(\epsilon/2)$-approximation of $f_i$ for $1 \le i \le n$. Assume that the optimal solution to the \rmt\, instance $(V, B, D, R, \widetilde{F})$ has a reward $J_R^{\dagger}$ and spends $t_1^{\dagger}, \ldots, t_n^{\dagger}$ time at the $n$ POI vertices. Using the approximation with the format given in~\eqref{equation:approximation-1} and satisfying~\eqref{equation:approximation-2}, we have
\begin{align}\label{equation:t11}
(1-\frac{\epsilon}{2}){f}_i(t_i) \le \widetilde{f}_i(t_i) \le (1+\frac{\epsilon}{2}){f}_i(t_i)
\end{align}
and
\begin{align}\label{equation:t12}
\frac{1}{1+\frac{\epsilon}{2}}\widetilde{f}_i(t_i) \le f_i(t_i) \le \frac{1}{1-\frac{\epsilon}{2}}\widetilde{f}_i(t_i).
\end{align}

Then (by~\eqref{equation:t12}), 
\begin{align}\label{equation:t13}
J_R^{\dagger} = \sum_{i = 1}^nf_i(t_i^{\dagger}) \le \frac{1}{1-\frac{\epsilon}{2}}\sum_{i = 1}^n\widetilde{f}_i(t_i^{\dagger}),
\end{align}
in which the summation $\sum_{i = 1}^n\widetilde{f}_i(t_i^{\dagger})$ is the reward returned by the MIP algorithm. On the other hand, by~\eqref{equation:t11}, we have
\begin{align}\label{equation:t14}
J_R^* = \sum_{i = 1}^n\widetilde{f}_i(t_i^*) \le (1+\frac{\epsilon}{2})\sum_{i = 1}^n{f}_i(t_i^*)
\end{align}
which implies that 
\begin{align}\label{equation:t15}
\sum_{i = 1}^n\widetilde{f}_i(t_i^{\dagger}) \le (1+\frac{\epsilon}{2})\sum_{i = 1}^n{f}_i(t_i^*).
\end{align}

To see that~\eqref{equation:t15} holds, assume instead it is false. Then, by by~\eqref{equation:t11} again, 
we have 
\[
(1+\frac{\epsilon}{2})\sum_{i = 1}^n{f}_i(t_i^*) < \sum_{i = 1}^n\widetilde{f}_i(t_i^{\dagger}) \le (1+\frac{\epsilon}{2})\sum_{i = 1}^n{f}_i(t_i^{\dagger}).
\]
We then have $J_R^* = \sum_{i = 1}^n{f}_i(t_i^*) < \sum_{i = 1}^n{f}_i(t_i^{\dagger})$, a contradiction. Putting~\eqref{equation:t13} and~\eqref{equation:t15} together yields
\[
J_R^{\dagger} \le \frac{1+\frac{\epsilon}{2}}{1-\frac{\epsilon}{2}}J_R^* = (1 + \epsilon + o(\epsilon))J_R^*.
\] ~\qed

For the \bmt\, problem, since time is split between traveling and actually staying at POIs, a direct $(1 + \epsilon)$-optimality assurance cannot be established. Nevertheless, for a \bmt\, instance with a reward requirement of $M_R > 0$, assuming that the optimal solution requires $J_T^*$ time, we can guarantee that a reward of at least $(1 - \epsilon)M_R$ is achieved using time no more than $J_T^*$. 

\begin{theorem}Given a \bmt\, instance specified by a 5-tuple $(V, B, D, R, F)$, $M_R > 0$, and a positive real number $\epsilon$, let its solution have a required total time of $J_T^*$. Then, an MIP model can be constructed that computes a solution with $J_R \ge (1 - \epsilon)M_R$ and $J_T \le J_T^*$. 
\end{theorem}
\noindent{\sc Proof.} For simplicity as well as diversity, we use a piece-wise linear approximation that is slightly different. Instead of making the piece-wise linear function satisfy~\eqref{equation:approximation-2}, we use only line segments that are no less $f_i$. That is, we can construct $\widetilde{f}_i$ such that 
\[
f_i(t_i) \le \widetilde{f}_i(t_i) \le \frac{1}{1 - \epsilon}f_i(t_i).
\]

Suppose that the optimal solution to the original \bmt\, instance spend $t_1^*, \ldots, t_n^*$ time at the $n$ POI vertices. Since for all $1 \le i \le n$, $\widetilde{f}_i(t_i) \ge f_i(t_i)$, the approximate MIP model constructed using $\widetilde{F}$ instead of $F$ will not need as much time to reach a reward of $M_R$. That is, the approximate model produces a solution with $J_T \le J_T^*$. Let the time spent at the POI vertices in the solution to the approximate MIP model be $t_1^{\dagger}, \ldots, t_n^{\dagger}$, then the actual achieved reward is
\[
J_R = \sum_{i = 1}^n f_i(t_i^{\dagger}) \ge \sum_{i = 1}^n (1 - \epsilon)\widetilde{f}_i(t_i^{\dagger}) = (1 - \epsilon)M_R.
\] ~\qed

\subsection{Extensions and Generalizations}\label{subsection:generalization}
Before concluding this section, we briefly mention a few extensions and generalizations of our MIP model. We only cover the \rmt\, problem here. The extension to \bmt\, is straightforward. 

\subsubsection*{Multiple tours} In a sense, the MIP model described so far creates a single-day itinerary since the plan is a single tour that starts and ends at the same base.  However, our MIP model can be easily generalized to allow the planning of trips with multiple tours. There are two possible generalizations with different applications. The first possibility is to force multiple tours to start at the same base, which represents the problem of a tourist staying at the same hotel for multiple days. Given the number of days $m$, we may obtain a more general MIP model by simply create $m$ copies of the edge-use variables, {\em i.e.}, $x_{ij}$'s. For each copy, a separate maximum time constraint (a daily time limit) is imposed. These $m$ copies are then aggregated together, {\em i.e.}, through $\sum_{k = 1}^m x_{ijk} = x_{ij}$.  We also require that the base POI vertices have either $0$ or $m$ incoming edges being used. the The rest of the MIP model remains essentially the same. 

The second possibility is applicable to multi-robot surveillance problems, in which all tours are disjoint. That is, each mobile robot covers a disjoint set of POIs to cooperatively collect the maximum amount of reward. Note that this implies $|B| \ge m$. To achieve this, we again create multiple copies of the edge-use variables, enforce the time constraints for each copy, and aggregate the variables. Then, we let the base POI vertices have at most a single incoming edge being used. 

\subsubsection*{Non-cyclic trip} The current MIP model forces a (cyclic) tour to be created. Whereas this may be more applicable to tourists, sometimes it may be beneficial to have non-cyclic routes. For example, using multiple hotels may allow a tourist to significantly increase the potential total reward due to reduced travel time. Alternatively, in a surveillance or monitoring problem, a single use probe may be sent to follow a one-time, non-cyclic route. To allow this, we may simply remove the constraint that forces both the incoming and outgoing edge from the origin vertex $o$ to a base vertex to be used. Non-cyclic trip can be directly combined with the multiple-tour generalization. 

\subsubsection*{Variations on learning curves} Although we focus on non-decreasing $C^1$ continuous learning curves with first order derivatives bounded away from zero, other types of learning curves can also be supported. The only requirement on the $f_i$'s is that they can be approximated arbitrarily well using a piece-wise linear, continuous function with a finite number of line segments. In particular, we note that the learning curve being non-monotone does not present an issue for the MIP model. Given a general $f_i$ that non-monotone, we can turn it into a non-decreasing function over which our MIP model can be applied. To do so, starting from $t_i = 0$, we find the first local maximum, say at $t_i = t_{i,1}$, at which point we augment $f_i$ by extending it from $f_i(t_{i,1})$ until it reaches the original $f_i$ at a point $t_i = t_{i, 2}$ where $f_i$ starts increasing again. We then repeat the same iterative process starting from $t_i = t_{i, 2}$. Such augmentation of $f_i$ is never problematic because our MIP model maximizes the reward using the least amount of time and will never allocate more time at a POI vertex when the reward is less or remains the same.


\section{The Algorithm and its Analysis}\label{section:algorithm}
\def\algo{{\sc OptimalTouristIntinerary}}
The overall algorithm construction is outlined in Algorithm~\ref{algorithm:all}. In Line~\ref{algorithm:fw} of the algorithm, it computes all-pairs shortest paths and their respective lengths using a transitive closure based algorithm, for example, the Floyd-Warshall algorithm \cite{Flo62,War62}. Then, in Lines \ref{algorithm:for}-\ref{algorithm:end-for}, the algorithm computes a piece-wise linear $(1+\epsilon/2)$-approximation of each $f_i \in F$, if necessary. Finally, Once $D'$ is computed and all of $\widetilde{F}$ is built, Lines~\ref{algorithm:split}-\ref{algorithm:end} of the algorithm can be carried out according to the steps outlined in Section~\ref{section:mip-model}. In the rest of this section, we cover two important properties of our algorithm. 

\begin{algorithm}
    \SetKwInOut{Input}{Input}
    \SetKwInOut{Output}{Output}
    \SetKwComment{Comment}{\%}{}
    \Input{$V$, the set of POIs, $B$, the set of bases, $D$, the set of (incomplete) inter-POI distances, $R$, the set of maximum POI rewards, $F$, the set of learning curves, $M_T$ (or $M_R$), the time (or reward) constraint, and	$\epsilon$, the required accuracy}
    \Output{$J_R^*$, the maximum attainable reward (or $J_T^*$, the minimum required time), and $E_{tr}$, a set of visited edges associated with $J_R^*$ (or $J_T^*$)}

\vspace{0.08in}
\Comment{\small Compute all pairs of shortest paths between all $1 \le i, j \le n$}
\vspace{0.025in}
$(P', D') \leftarrow$ {\sc FloydWarshall}$(V, D)$ \label{algorithm:fw}

\vspace{0.08in}
\Comment{\small Compute for each $f_i \in F, 1 \le i \le n$, a piece-wise linear $(1+\epsilon/2)$-approximation}
\vspace{0.025in}
\For{$f_i \in F, 1 \le i \le n$\label{algorithm:for}}{
$\widetilde{f}_i \leftarrow$ {\sc ComputeEpsilonApproximation}$(f_i, \epsilon/2)$
}\label{algorithm:end-for}

\vspace{0.08in}
\Comment{\small Setting up the MIP model and optimize it using an MIP solver}
\vspace{0.025in}
$(V', D') \leftarrow $ {\sc VertexSplit}$(V, B, D')$ \Comment*{\small Split $v \in B$}\label{algorithm:split}
{\sc BuildModel}$(V', D, R, \widetilde{F})$ \Comment*{\small Also builds $J_R$ and $J_T$} 

\uIf{$M_T$ is given}{Set $J_T \le M_T$ and maximize $J_R$ \Comment*{\small Maximize reward} }
\Else{Set $J_R \ge M_R$ and minimize $J_T$ \Comment*{\small Minimize time} }\label{algorithm:end}

\vspace{0.08in}
\Return{$J_R^*$ (or $J_T^*$), and the associated $E_{tr}$}
\caption{\algo} \label{algorithm:all}
\end{algorithm}

\subsection{Finite Complexity of Piece-Wise Linear Approximation}
In Section~\ref{section:mip-model}, we mentioned that a reasonably nice learning curve can be approximated to arbitrary precision using a piece-wise linear function, which is not difficult to imagine. However, to encode the approximated piece-wise linear function into the MIP model, the function must have finitely many line segments. We now show that the approximation indeed has limited complexity.

\begin{figure}[htp]
\begin{center}
    \includegraphics[width=3.15in]{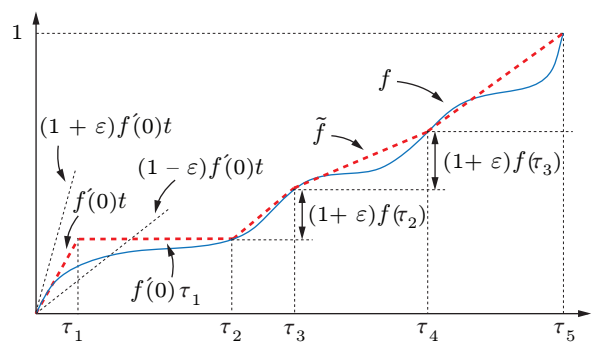} 
\end{center}
\caption{\label{figure:finite-proof} A graphical illustration of the constructive proof for Theorem~\ref{theorem:approximation}.}
\end{figure}

\begin{theorem}\label{theorem:approximation}Let $f \in [0, 1]$ be a $C^1$ continuous, non-decreasing function with $f(0) = 0$ and $f'(0) \ge \lambda$ for some fixed $\lambda > 0$. For any given $\epsilon > 0$, there exists a piece-wise linear approximation of $f$ containing only finite number of line segments, denoted $\widetilde f$, such that 
\begin{align}\label{equation:t3}
\frac{\vert f(t) - \widetilde{f}(t) \vert}{f(t)} \le \epsilon. 
\end{align}
\end{theorem}
\noindent{\sc Proof.} At $t = 0$, by the continuity of $f'(t)$, for an arbitrary $\lambda\epsilon > 0$, there exists $t_{\delta}$ such that for all $0 \le t \le t_{\delta}$,
\begin{align}\label{equation:t31}
f'(0) - \lambda\epsilon \le f'(t) \le f'(0) + \lambda\epsilon. 
\end{align}
Since $f'(0) \ge \lambda$, we obtain from~\eqref{equation:t31} that
\begin{align}\label{equation:t32}
(1 - \epsilon)f'(0) \le f'(t) \le (1 + \epsilon)f'(0). 
\end{align}
Then, since $f(0) = 0$,~\eqref{equation:t32} implies that 
\begin{align}\label{equation:t33}
(1 - \epsilon)f'(0)t \le f(t) \le (1 + \epsilon)f'(0)t. 
\end{align}
We let the first (left most) line segment of the approximation $\widetilde{f}$ be simply $f'(0)t$ for $0 \le t \le t_{\delta} =: \tau_1$ (see Figure~\ref{figure:finite-proof} for a graphical illustration). Then, the second inequality of~\eqref{equation:t33} becomes
\[
f(t) \le (1+\epsilon)\widetilde{f}(t) \Rightarrow \frac{1}{1 + \epsilon}f(t) \le \widetilde{f}(t) \Rightarrow (1-\epsilon)f(x) \le \widetilde{f}(t), 
\]
which implies~\eqref{equation:t3} for $0 \le t \le \tau_1$. Same holds for the first inequality of~\eqref{equation:t33}. 

For the second line segment, we simply extend from $(\tau_1, f'(0)\tau_1)$ either horizontally (when $f'(0)\tau_1 > f(\tau_1)$ or vertically  (when $f'(0)\tau_1 < f(\tau_1)$ until the line segment meets $f$. Let this point on $f(t)$ be $(\tau_2, f(\tau_2))$. 

The rest of $\widetilde{f}$ can then be iteratively defined starting from the point $(\tau_2, f(\tau_2))$. For the third line segment, we let its end point be $(\tau_3, f(\tau_3))$ such that $f(\tau_3) = \min\{1, (1+\epsilon)f(\tau_2)\}$. Because $f$ is non-decreasing, over $\tau_2 \le t \le \tau_3$, 
\[
f(\tau_2) \le f(t) \le f(\tau_3) \le (1 + \epsilon)f(\tau_2),
\]
the same holds true for $\widetilde{f}$ over $\tau_2 \le t \le \tau_3$. Therefore, over $\tau_2 \le t \le \tau_3$,
\[
\vert \widetilde{f}(t) - f(t) \vert \le \epsilon f(\tau_2) \Rightarrow \frac{\vert \widetilde{f}(t) - f(t) \vert}{f}(t) \le \epsilon\frac{f(\tau_2)}{f(t)} \le \epsilon.
\]
We can then iteratively define the rest of $\widetilde{f}$ similarly. Because each time we extend $\widetilde{f}$ by $(1+ \epsilon)$ and we start from $f(\tau_2) > 0$, in finite number of iterations $\widetilde{f}$ reaches 1. ~\qed

\begin{figure}[htp]
\begin{center}
    \includegraphics[width=3.15in]{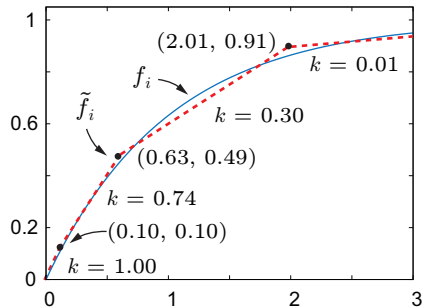} 
\end{center}
\caption{\label{figure:exponential} A graphical illustration of the constructive proof for Theorem~\ref{theorem:approximation}. The different $k$'s indicate the slopes of the corresponding line segments.}
\end{figure}

\textbf{Remark.} We emphasize that the constructive proof of Theorem~\ref{theorem:approximation} may yield approximations that are far from the best piece-wise linear approximations. On the other hand, practical, non-linear learning functions often do not require complex piece-wise linear functions to approximate. As an example, when a learning curve from the exponential family is used, {\em e.g.}, $f_i(t_i) = 1 - e^{-\lambda_i t_i}$, a $1.05$-approximation of $f_i$ can be achieved using only four line segments. Since the derivative of $f_i$ can be easily computed in this case, numerically computing the approximation is fairly easy. Moreover, only a one-time computation is required; simple scaling can then extend the computation easily to different learning rates ($\lambda_i$'s) and rewards ($r_i$'s). The initial and the approximated curves for the case of $\lambda_i = 1$ are illustrated in Figure~\ref{figure:exponential}. It is straightforward to verify that $\vert\widetilde{f}_i - f_i\vert/f_i < 0.05$. 

\subsection{The Anytime Property}
An very useful property of Algorithm~\ref{algorithm:all} that we obtain for free is that it yields an {\em anytime} algorithm. The anytime property is a direct consequence of solving the MIP models for \rmt\, and \bmt\, using an MIP solver, which generally use some variations of the branch-and-bound algorithm \cite{LanDoi60}. Roughly speaking, a branch-and-bound algorithm works with a (high-dimensional) polytope that contains all the feasible solutions to an optimization problem. The algorithm then iteratively partitions the polytope into smaller ones and truncates more and more of the polytope that are known not to contain the optimal solution. After some initial steps, a tree structure is built and the leaves of the tree contains portions of the original feasibility polytope that are still active. For each of these polytopes, suppose we are working on a maximization problem, it is relatively easy to locate a feasible solution with the correct integrality condition ({\em i.e.}, a feasible solution in which binary/integer variables get assigned binary/integer values). The maximum of all these feasible solution is then a lower bound of the optimal value. On the other hand, it is also possible to compute for each leaf the maximum achievable objective without respecting the integrality constraints, which yields a lower bound on the optimal value. The difference between the two bounds is often referred to as the {\em gap}. When the gap is zero, the optimal solution is found. Over the running course of a branch-and-bound algorithm, if the gap gradually decreases, an anytime algorithm is obtained. 

For our particular problems, the anytime property is quite useful since computing the true optimal solution to the (potentially approximate) MIP model for \rmt\, and \bmt\, can be very time consuming. We will see in Section~\ref{section:experiment} that for medium sized problems, a 1.2-optimal solution, which is fairly good for practical purposes, can often be computed quickly. 
\section{Computational Experiments}\label{section:experiment}
In this section, we evaluate our proposed algorithm in several computational experiments. In these experiments, we look at the solution structure, computational performance, and an application to planning a day tour of Istanbul. The simulation is implemented in the Java programming language. For the MIP solver, Gurobi \cite{gurobi} is used. Our computational experiments were carried out on an Intel Core-i7 3930K PC with 64GB of memory. 

\subsection{Anytime Solution Structure}\label{subsection:anytime}
\begin{figure}[htp]
\begin{center}
\begin{tabular}{cc}
    \includegraphics[width=1.6in]{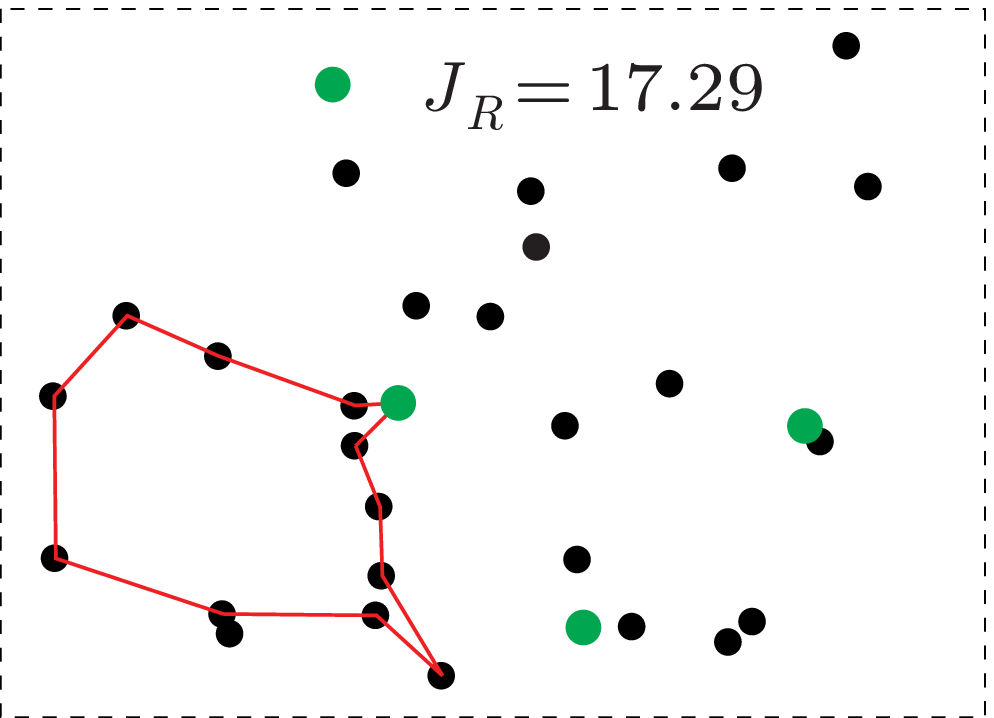} &
    \includegraphics[width=1.6in]{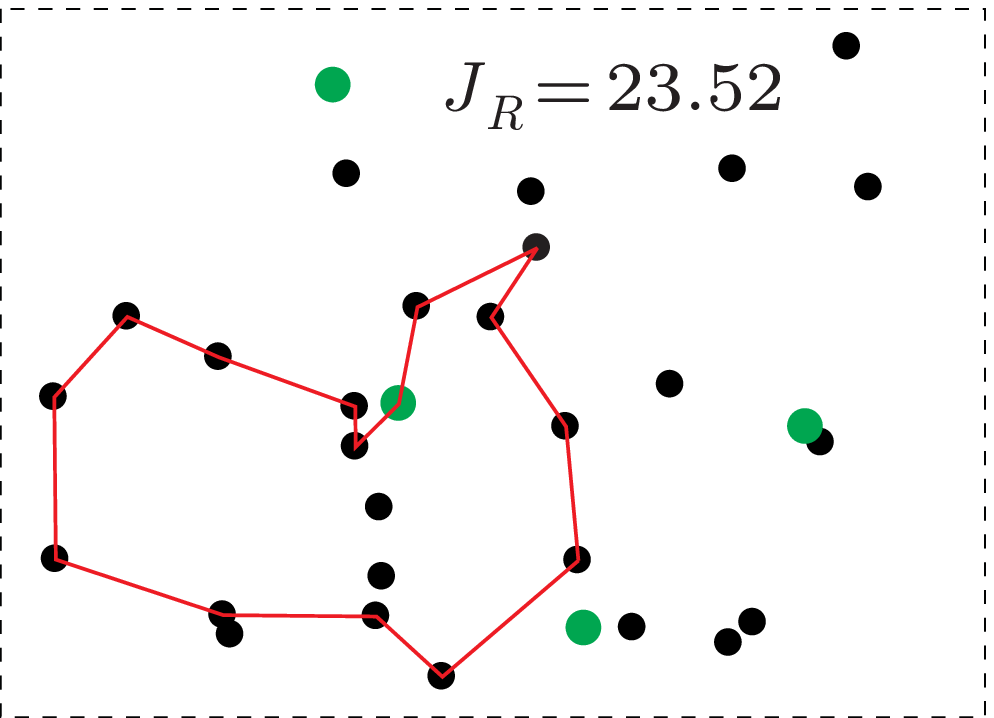} \\
		(a) & (b) \\
    \includegraphics[width=1.6in]{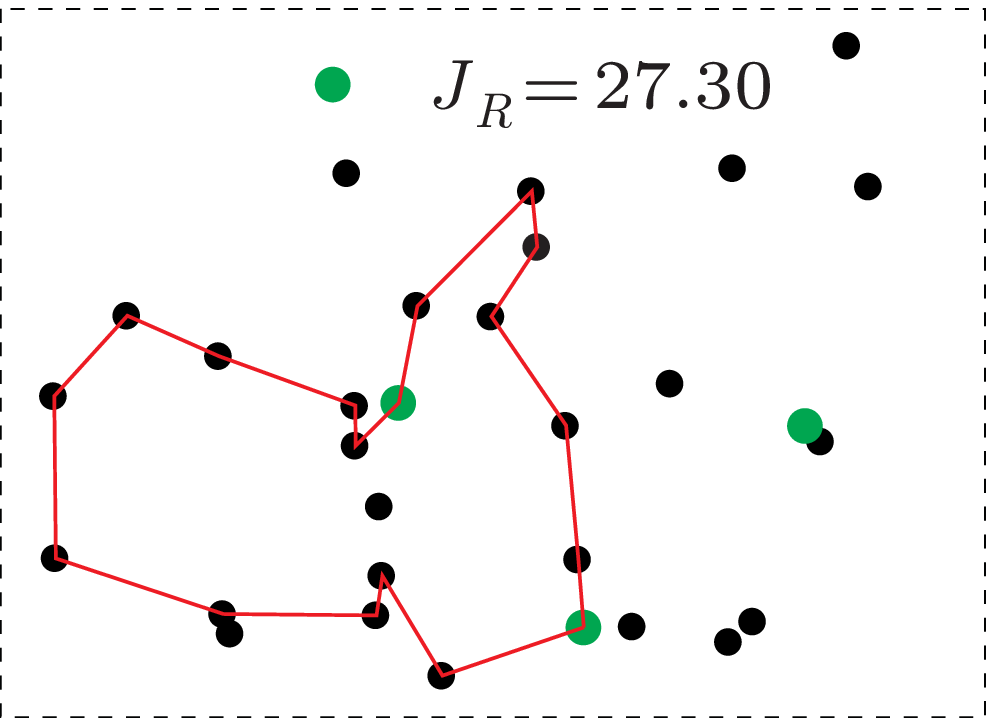} &
    \includegraphics[width=1.6in]{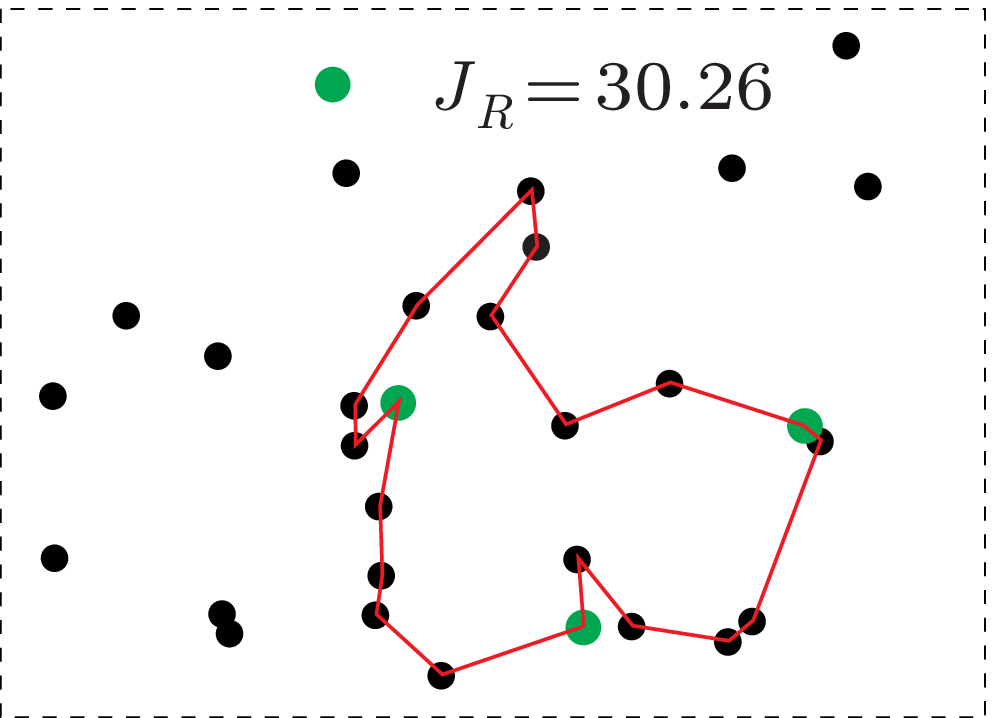} \\
		(c) & (d) \\
    \includegraphics[width=1.6in]{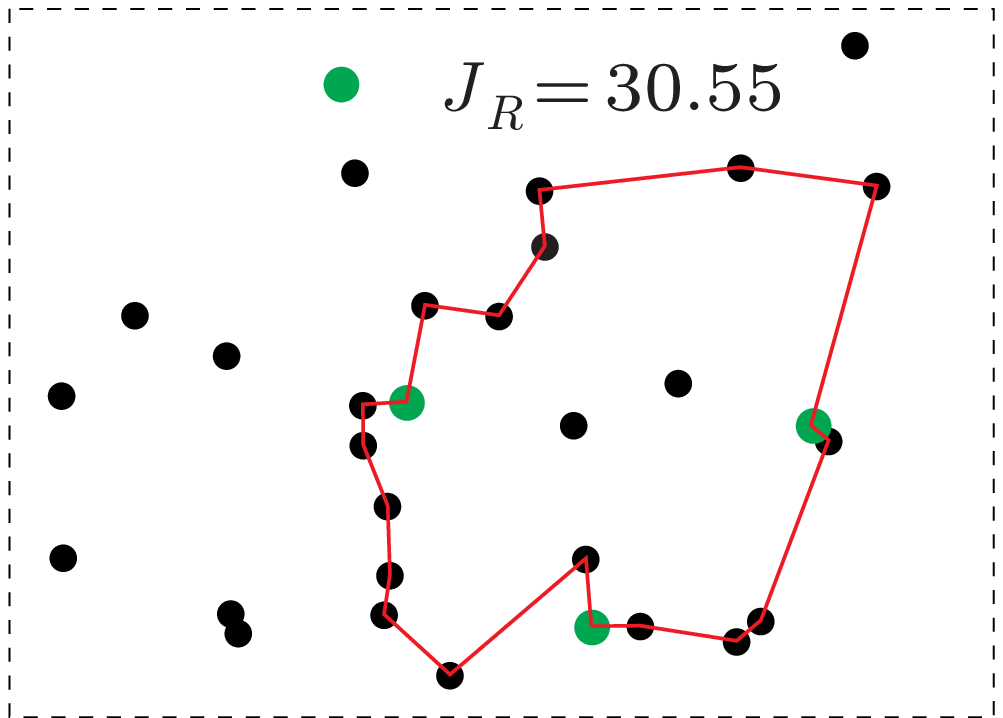} & \\
		(e) & \\
\end{tabular}
\end{center}
\caption{\label{figure:rmt-linear} Figures (a) - (e): POIs visited by the best solution to the  \rmt\, problem after the gap dips just below $100\%$, $50\%$, $20\%$, $10\%$, and $5\%$, respectively. The solution obtained after the gap dips below $5\%$ is in fact the optimal solution for this particular example. The black and the green dots are the POIs and the green dots are the base vertices. }
\end{figure}

Our first set of experiments was performed over a randomly generated example, created in the following way. The example contains $30$ uniformly randomly distributed POIs in a $10 \times 15$ rectangle (see Figure~\ref{figure:rmt-linear}). Each POI $v_i$ is associated with a $\lambda_i \in [1, 2)$ and an $r_i \in [1, 2)$ that were both uniformly randomly selected. The $\lambda_i$'s and $r_i$'s are selected not to vary by much because we expect that in practice, this will present a more difficult choice for a tourist or a mobile robot. 
For $f_i$, both linear ({\em e.g.}, with the form~\eqref{equation:linear}) and exponential ({\em e.g.}, with the form~\eqref{equation:exponential}) types were used, with the learning rates specified by the $\lambda_i$'s. We set $\epsilon = 0.05$ when we approximate the non-linear $f_i$'s with piece-wise linear functions (that is, we use the linear approximation illustrated in Figure~\ref{figure:exponential} with proper scaling). Note that $\epsilon = 0.05$ yields a 1.1-optimal MIP model for exponential $f_i$'s. These steps determine the sets $V$, $R$, $F$, $\widetilde{F}$. We let $B$ to be the set $\{v_1, v_9, v_{17}, v_{25}\}$. For deciding $E$ and $D$, we let there be an edge between two POI vertices $v_i, v_j$ if the Euclidean distance between them is no more than $10$. Finally, the constraints were set as follows. For \rmt, $M_T = 50$ for both linear and exponential $f_i$'s. For \bmt, $M_R = 30.55$ for linear $f_i$'s and $M_R = 25.78$ for exponential $f_i$'s. These $M_R$'s were selected because they are the optimal $J_R$ value for the respective \rmt\, problems with $M_T = 50$. 

For each problem instance, we extract the solution after the gap becomes no more than $100\%$, $50\%$, $20\%$, $10\%$, $5\%$, $1\%$, and $0\%$. These solutions for the \rmt\, instance with linear learning curves are illustrated in Figure~\ref{figure:rmt-linear}. Because the large number of POIs involved, we do not list the computed $t_i$'s but point out that, in the linear case, when the set of POIs for staying is selected, it is always beneficial to exhaust the reward at POIs with the largest learning rate since time is best used this way. The computation of these five solutions took $0.96, 1.05, 2.16, 3.70$, and $10.2$ seconds, respectively. Confirming that the last solution (Figure~\ref{figure:rmt-linear}(e)) is indeed the optimal solution took $76$ seconds. 

\begin{figure}[htp]
\begin{center}
\begin{tabular}{cc}
    \includegraphics[width=1.6in]{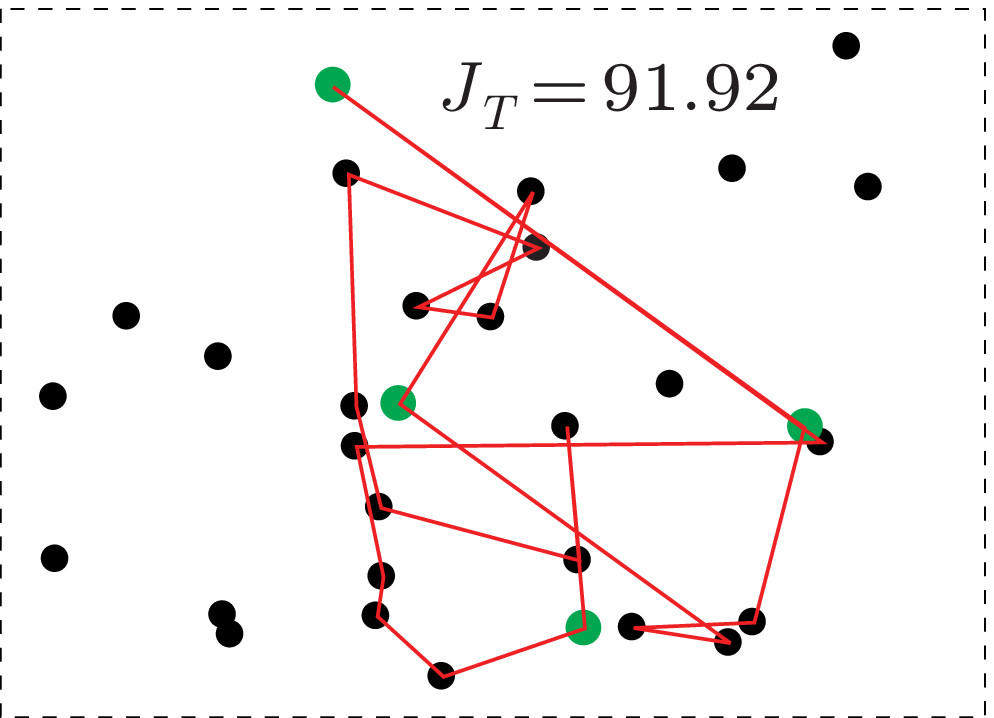} &
    \includegraphics[width=1.6in]{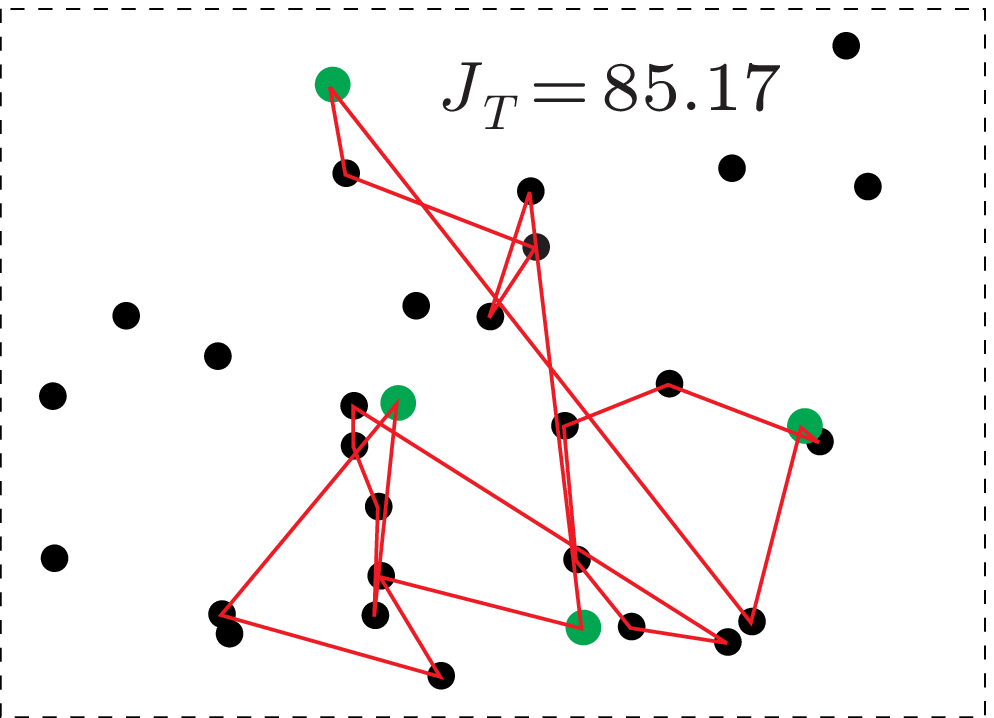} \\
		(a) & (b) \\
    \includegraphics[width=1.6in]{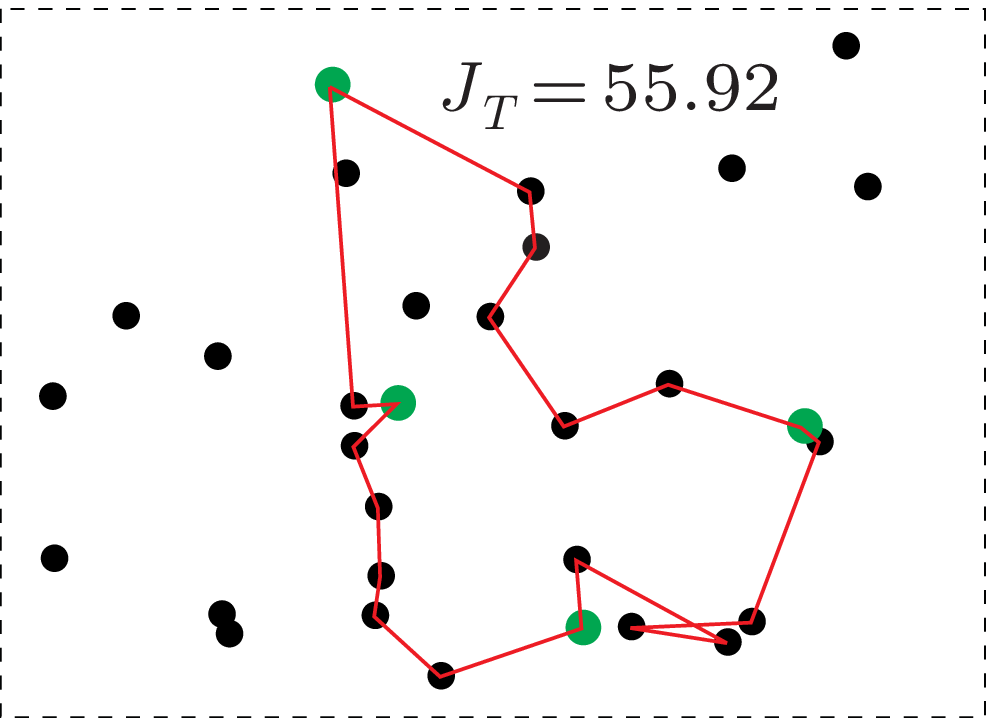} &
    \includegraphics[width=1.6in]{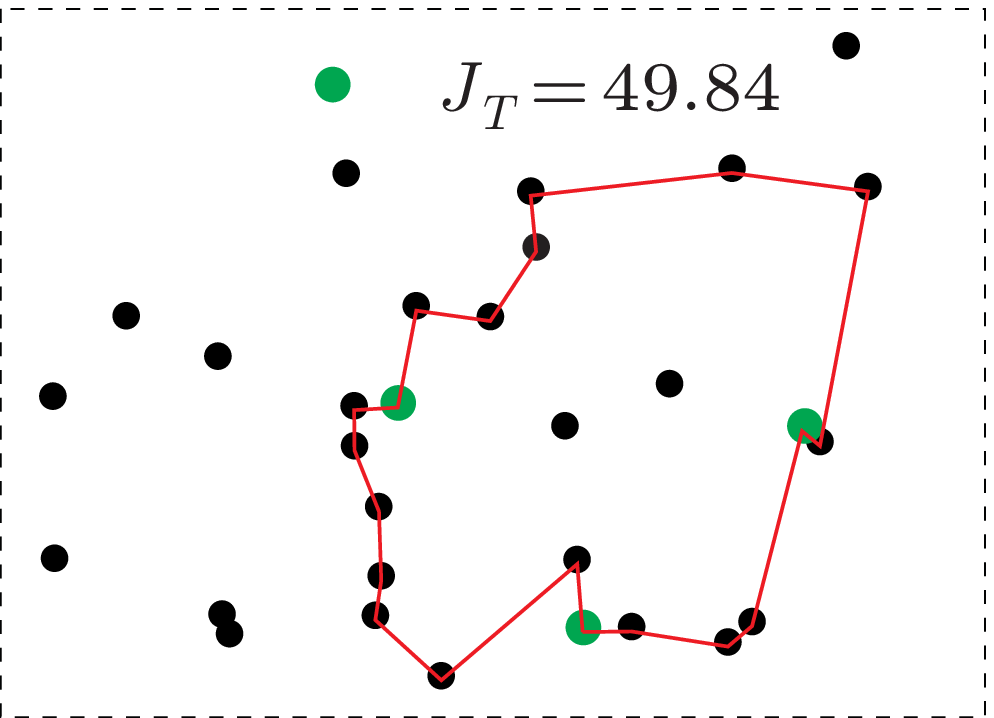} \\
		(c) & (d) \\
    \includegraphics[width=1.6in]{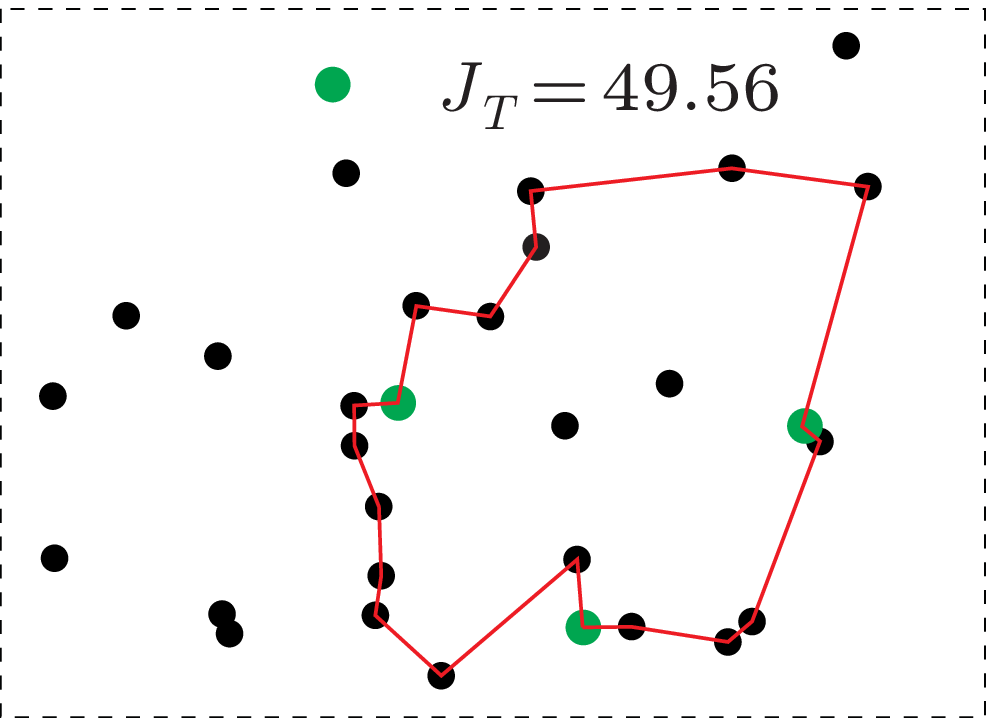} &
     \\
		(e) &
\end{tabular}
\end{center}
\caption{\label{figure:bmt-linear} Figures (a) - (e): POIs visited by the best solution to the \bmt\, problem after the gap dips just below $100\%$, $50\%$, $20\%$, $10\%$, and $1\%$, respectively.}
\end{figure}

For the \bmt\, instance with $M_R = 30.55$, we similarly plot the solutions at different accuracies in Figure~\ref{figure:bmt-linear}. Note that the optimal solution (Figure~\ref{figure:bmt-linear}(e)) yields the same tour as the optimal solution to the corresponding \rmt\, problem (Figure~\ref{figure:rmt-linear}(e)). We note that $J_T$ is actually smaller than $50$ in this case, suggesting $M_T = 50$ is not necessary to reach a reward of $J_R = 30.55$. The computation of these five solutions took $0.48, 0.60, 3.13, 6.19$, and $28.60$ seconds, respectively. Confirming that the last solution is indeed the optimal solution took $50$ seconds. 

\begin{figure}[htp]
\begin{center}
\begin{tabular}{cc}
    \includegraphics[width=1.6in]{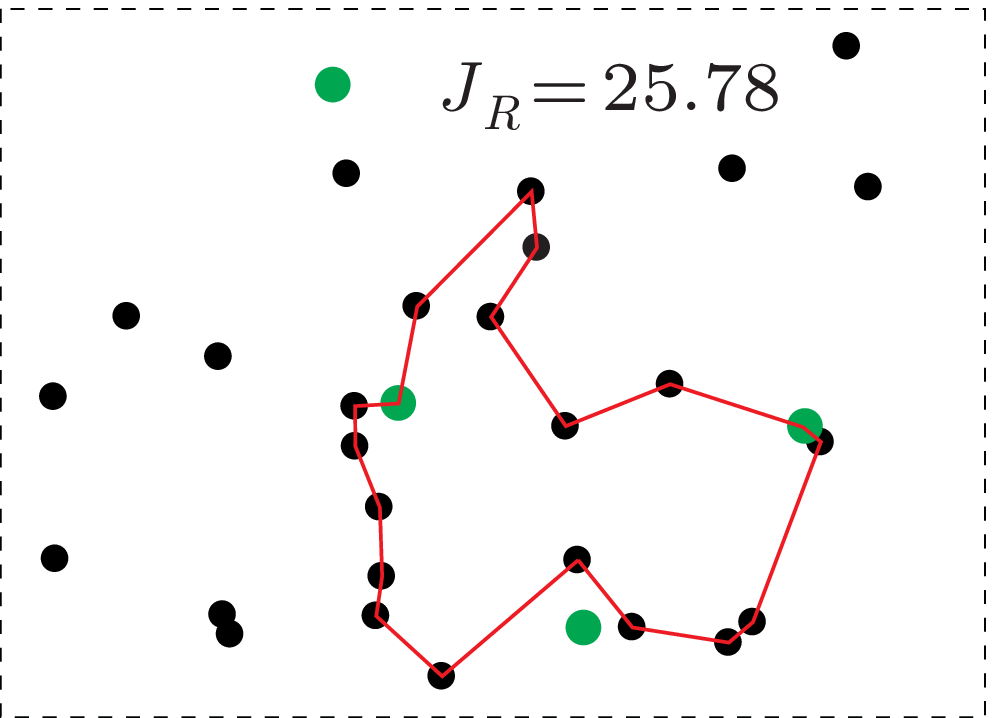} &
    \includegraphics[width=1.6in]{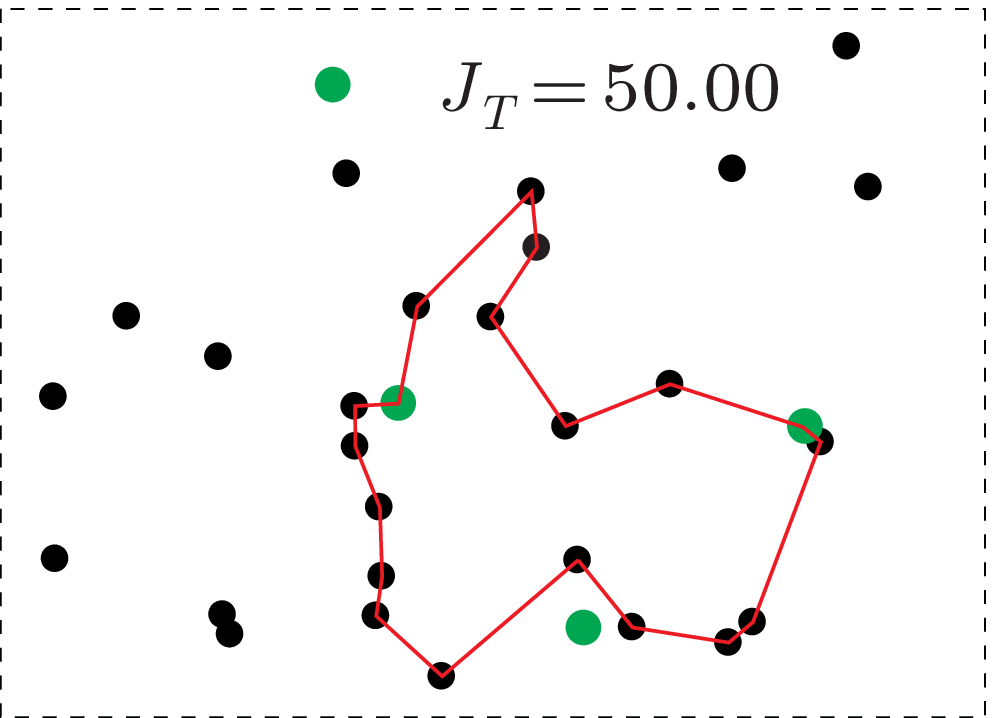} \\
		(a) & (b) 
\end{tabular}
\end{center}
\caption{\label{figure:rmt-bmt-exp} (a) Optimal solution to \rmt\, with exponential learning curves and $M_T = 50.00$. (b) Optimal solution to \bmt\, with exponential learning curves with $M_R = 25.78$.}
\end{figure}

For exponential learning curves, similar results were obtained. The optimal tours for \rmt\, and \bmt\, are illustrated in Figure~\ref{figure:rmt-bmt-exp}, which, as expected, have the same tour. Computing the optimal solution to these more complex 1.1-optimal MIP models took $27.6$ and $30.1$ seconds, respectively. 

\subsection{Computational Performance}
Since the models for \rmt\, and \bmt\, attempt to solve an NP-hard problem precisely (note that the problem after linearization remains NP-hard), no polynomial time algorithm exists unless P = NP. Therefore, our evaluation of the algorithm's computational performance is limited to an empirical one. For this, two large sets of computations are performed. In the first set of computations, rectangular grids of various sizes were constructed. The POIs reside on the lattice points on these grid, with the reward and learning rate selected uniformly randomly from $[1, 2)$. Vertices $n/3$ and $2n/3$ are selected as base vertices.  For each choice of grid sizes, 10 example problems are created. For the \rmt\, instances, a time budget of 1.5 times the grid perimeter is used. For the \bmt\, instances, a reward requirement of 0.6 times the grid perimeter is used. These constraints are chosen to allow the tour to go through $10\%$ to $25\%$ of the total POIs. For both \rmt\, and \bmt\, instances, we perform computations with both linear and exponential learning curves (with $5\%$ linearization). The average time, in seconds, required to compute a solution up to given accuracy is listed in Table~\ref{table:performance-grid}. The number in the parenthesis denote the number of times, out of a total of ten, that the computation completed within a limit of 900 seconds. 

\begin{table*}[t]
\begin{center}
	 \caption{\label{table:performance-grid}Computation time for solving \rmt\, and \bmt\, over POIs located at the lattice points on various sized integer grids.}
	 \begin{tabularx}{\textwidth}{cccXXXXXXX}
   \hline\hline
	 \multirow{2}*{grid size} & \multirow{2}*{problem}  & \multirow{2}*{learning curve} & \multicolumn{7}{c}{MIP gap} \\
	 \cline{4-10} & & & $100\%$ & $50\%$ & $20\%$ & $10\%$ & $5\%$ & $1\%$ & $0\%$\\
	 \hline
  	 \multirow{4}*{ $4 \times 5$} 
& \rmt\, &linear & 0.085s (10) & 0.135s (10) & 0.203s (10) & 0.261s (10) & 0.675s (10) & 2.285s (10) & 2.357s (10) \\ 
& \bmt\, &linear & 0.070s (10) & 0.108s (10) & 0.271s (10) & 0.571s (10) & 0.974s (10) & 1.101s (10) & 1.102s (10) \\ 
& \rmt\, &exponential & 0.149s (10) & 0.171s (10) & 0.240s (10) & 0.388s (10) & 0.471s (10) & 1.293s (10) & 1.343s (10) \\ 
& \bmt\, &exponential & 0.061s (10) & 0.090s (10) & 0.174s (10) & 0.364s (10) & 0.505s (10) & 0.605s (10) & 0.608s (10) \\ 
	 \hline
	 \multirow{4}*{ $5 \times 6$} 
& \rmt\, &linear & 0.309s (10) & 0.342s (10) & 0.439s (10) & 0.531s (10) & 1.561s (10) & 17.00s (10) & 18.66s (10) \\ 
& \bmt\, &linear & 0.191s (10) & 0.250s (10) & 0.868s (10) & 2.038s (10) & 5.580s (10) & 8.038s (10) & 8.080s (10) \\ 
& \rmt\, &exponential & 0.361s (10) & 0.395s (10) & 0.522s (10) & 0.814s (10) & 1.225s (10) & 11.31s (10) & 12.41s (10) \\ 
& \bmt\, &exponential & 0.147s (10) & 0.194s (10) & 0.586s (10) & 1.803s (10) & 5.710s (10) & 9.383s (10) & 9.483s (10) \\ 
	 \hline
	 \multirow{4}*{ $6 \times 7$} 
& \rmt\, &linear & 0.683s (10) & 0.687s (10) & 0.816s (10) & 1.009s (10) & 5.790s (10) & 161.3s (7) & 209.8s (7) \\ 
& \bmt\, &linear & 0.501s (10) & 0.514s (10) & 6.308s (10) & 31.76s (10) & 79.22s (10) & 127.9s (10) & 129.0s (10) \\ 
& \rmt\, &exponential & 0.870s (10) & 0.914s (10) & 1.784s (10) & 5.268s (10) & 17.91s (10) & 182.6s (8) & 234.6s (8) \\ 
& \bmt\, &exponential & 0.701s (10) & 0.715s (10) & 3.718s (10) & 11.37s (10) & 78.40s (10) & 79.43s (8) & 80.87s (8) \\ 
	 \hline
	 \multirow{4}*{ $8 \times 10$} 
& \rmt\, &linear & 2.272s (10) & 2.443s (10) & 2.953s (10) & 21.58s (10) & 87.13s (10) & 454.6s (3) & 809.0s (1) \\ 
& \bmt\, &linear & 2.188s (10) & 2.382s (10) & 3.111s (10) & 20.75s (10) & 134.1s (9) & 284.2s (6) & 295.2s (6) \\ 
& \rmt\, &exponential & 2.134s (10) & 2.345s (10) & 5.664s (10) & 22.75s (10) & 67.28s (10) & 342.5s (4) & 498.5s (3) \\ 
& \bmt\, &exponential & 2.530s (10) & 2.849s (10) & 20.64s (10) & 79.32s (10) & 274.6s (9) & 492.9s (6) & 524.7s (6) \\ 
	 \hline
	 \multirow{4}*{ $10 \times 20$} 
& \rmt\, &linear & 17.31s (10) & 17.31s (10) & 18.96s (10) & 98.66s (10) & 433.9s (7) & N/A     & N/A     \\ 
& \bmt\, &linear & 43.28s (10) & 48.84s (10) & 93.40s (10) & 241.9s (9) & 346.8s (4) & N/A     & N/A     \\ 
& \rmt\, &exponential & 17.33s (10) & 26.87s (10) & 48.06s (9) & 59.64s (5) & 317.3s (1) & N/A     & N/A     \\ 
& \bmt\, &exponential & 37.17s (10) & 44.29s (10) & 241.3s (10) & 424.2s (6) & 435.4s (1) & N/A     & N/A     \\ 
	 \hline\hline
	 \end{tabularx}
\end{center}
\end{table*}

Our second set of computations generates the POI locations uniformly randomly according to the same rules used in Section~\ref{subsection:anytime}, in a $\vert V\vert \times 1.2\vert V\vert$ rectangle. Then, for \rmt\, instances, a time budget of $4\sqrt{\vert V\vert}$ is used. For \bmt\, instances, a reward requirement of $2\sqrt{\vert V\vert}$ is used. The rest of the setup is done similarly as in the rectangular grid case. The computational performance is listed in Table~\ref{table:performance-random}.

\begin{table*}[t]
\begin{center}
	 \caption{\label{table:performance-random}Computation time for solving \rmt\, and \bmt\, over POIs that are uniformly randomly selected.}
	 \begin{tabularx}{\textwidth}{cccXXXXXXX}
   \hline\hline
	 \multirow{2}*{$\#$ of samples} & \multirow{2}*{problem}  & \multirow{2}*{learning curve} & \multicolumn{7}{c}{MIP gap} \\
	 \cline{4-10} & & & $100\%$ & $50\%$ & $20\%$ & $10\%$ & $5\%$ & $1\%$ & $0\%$\\
	 \hline
\multirow{4}*{ $20$} 
& \rmt\, & linear& 0.118s (10)& 0.236s (10)& 0.730s (10)& 2.645s (10)& 4.832s (10)& 5.887s (10)& 5.92 s (10)\\ 
& \bmt\, & linear& 0.049s (10)& 0.112s (10)& 1.183s (10)& 1.727s (10)& 1.908s (10)& 1.962s (10)& 1.966s (10)\\ 
& \rmt\, & exponential& 0.274s (10)& 0.379s (10)& 3.780s (10)& 6.499s (10)& 13.38s (10)& 17.49s (10)& 17.57s (10)\\ 
& \bmt\, & exponential& 0.071s (10)& 0.166s (10)& 2.531s (10)& 5.731s (10)& 6.946s (10)& 7.400s (10)& 7.424s (10)\\ 
 	 \hline
 	 \multirow{4}*{ $30$} 
& \rmt\, & linear& 0.435s (10)& 1.122s (10)& 15.41s (10)& 74.97s (10)& 228.1s (9)& 81.90s (7)& 82.95s (7)\\ 
& \bmt\, & linear& 0.289s (10)& 0.821s (10)& 18.23s (10)& 54.56s (10)& 76.39s (10)& 81.34s (10)& 81.58s (10)\\ 
& \rmt\, & exponential& 1.221s (10)& 2.664s (10)& 17.44s (10)& 81.39s (10)& 168.4s (9)& 159.2s (8)& 161.8s (8)\\ 
& \bmt\, & exponential& 0.339s (10)& 1.243s (10)& 8.676s (10)& 24.96s (10)& 41.74s (10)& 47.93s (10)& 48.15s (10)\\ 
 	 \hline
 	 \multirow{4}*{ $42$} 
& \rmt\, & linear& 6.058s (10)& 13.73s (10)& 49.33s (8)& 164.5s (6)& 262.9s (3)& 170.2s (1)& 187.8s (1)\\ 
& \bmt\, & linear& 0.612s (10)& 1.513s (10)& 142.2s (9)& 107.0s (7)& 141.2s (7)& 148.9s (7)& 149.1s (7)\\ 
& \rmt\, & exponential& 14.13s (10)& 24.80s (10)& 93.65s (10)& 132.2s (6)& 288.2s (4)& 375.9s (3)& 381.1s (3)\\ 
& \bmt\, & exponential& 1.711s (10)& 7.759s (10)& 179.4s (7)& 195.4s (4)& 279.6s (3)& 362.7s (3)& 365.0s (3)\\ 
 	 \hline
 	 \multirow{4}*{ $100$} 
& \rmt\, & linear& 54.26s (10)& 57.26s (10)& 439.3s (8)& N/A    & N/A    & N/A    & N/A    \\ 
& \bmt\, & linear& 19.98s (10)& 105.5s (10)& N/A    & N/A    & N/A    & N/A    & N/A    \\ 
& \rmt\, & exponential& 59.45s (10)& 125.3s (9)& 309.0s (5)& 790.8s (1)& N/A    & N/A    & N/A    \\ 
& \bmt\, & exponential& 12.03s (10)& 251.1s (10)& 577.3s (1)& N/A    & N/A    & N/A    & N/A    \\ 
 	 \hline
 	 \multirow{4}*{ $200$} 
& \rmt\, & linear& 40.26s (10)& 255.0s (8)& N/A    & N/A    & N/A    & N/A    & N/A    \\ 
& \bmt\, & linear& 170.7s (10)& 185.1s (9)& N/A    & N/A    & N/A    & N/A    & N/A    \\ 
& \rmt\, & exponential& 34.50s (10)& 229.5s (8)& N/A    & N/A    & N/A    & N/A    & N/A    \\ 
& \bmt\, & exponential& 223.4s (10)& 494.3s (3)& N/A    & N/A    & N/A    & N/A    & N/A    \\ 
	 \hline\hline
	 \end{tabularx}
\end{center}
\end{table*}

From the computational experiments, we observe that in the grid case, for up to $200$ POIs, the proposed method can compute a $1.2$-optimal (corresponding to a $20\%$ gap) MIP solution for almost all instances ($199$ out of $200$ instances), under very reasonable computation time. Moreover, for up to $80$ POIs, the method can compute a $1.05$- optimal MIP solution for almost all instances ($158$ out of $160$ instances). When the POIs are selected randomly, the computation seems to be more challenging. Computing $1.2$-optimal MIP solution starts to become challenging when there are more than $40$ POIs. The difficulty seems to come from the fact that randomly selected POIs can potentially be packed more densely in certain local regions. Nevertheless, we were still able to compute $1.5$-optimal MIP solutions in most of the cases when there are 100 POIs. Overall, the two large sets of computations suggest that our algorithm can be used to do itinerary planning for practical-sized instances in large cities.  

\subsection{Planning a One-Day Istanbul Tour}
As a last computational example, we illustrate how one may use real data to compute a day tour of Istanbul over 20 POIs.\footnote{We intentionally limited the size and complexity of this example to provide all important details.} These 20 POIs are selected by taking the top-ranked attractions from TripAdvisor's\footnote{\texttt{http://www.tripadvisor.com}} city guide for Istanbul. We select the top 20 POIs that are not general areas and have at least 300 user reviews. These POIs are (the ordering is by the POI's rank): 1. Suleymaniye Mosque, 2. Rahmi M. Koc Museum, 3. Rustem Pasha Mosque, 4. Hagia Sophia Museum, 5. Kariye Museum, 6. Basilica Cistern, 7. Bosphorus Strait, 8. Blue Mosque, 9. Rumeli Fortress, 10. Eyup Sultan Mosque, 11. Kucuk Ayasofya Camii, 12. Topkapi Palace, 13. Miniaturk, 14. Istanbul Archaeological Museums, 15. Gulhane Park, 16. Istanbul Modern Museum, 17. New Mosque, 18. Dolmabahce Palace, 19. The Bosphorus Bridge, and 20. Galata Tower. 

After the POIs are selected, we compute the maximum reward of these POIs using the formula $\sqrt[3]{n_{review}} + 10 - rank/5$, in which $n_{review}$ is the total number of reviews received for the POI on TripAdvisor and $rank$ is the POI's rank on TripAdvisor. The attractions are mostly museums and architectural sites, to which we assign the learning rates of $1 - 0.01r_i$, {\em i.e.}, we expect a tourist to spend more time at more renowned POIs . Using Google Map\footnote{\texttt{http://maps.google.com}.}, we extracted the pair-wise distances between any two of these POIs and build the sets $E$ and $D$. The base vertex set is selected to contain the 1st, 6th, 11th, and 16th ranked POIs. With these parameters, we solve the \rmt\, problem with exponential learning curves and a time budget of 9 hours. From the solution (an exact solution to the $1.1$-optimal MIP model, computed in about five seconds) we extracted the itinerary listed in Table~\ref{table:itinerary}. The itinerary visits 14 POIs and yields a reward of 115 out of a total possible reward of 380. A visual inspection of the itinerary suggests that it is a fairly reasonable solution to our proposed problem. 

\begin{table}[htp]
\begin{center}
\caption{\label{table:itinerary} A 9-hour computed itinerary in Istanbul. }
	\begin{tabularx}{\columnwidth}{cX}
	 \hline\hline	 \\
	 1 & Start from the Suleymaniye Mosque, stay for 0.84 hour  \\
	 2 & Take a taxi to Topkapi Palace (8 min), stay for 0.88 hour  \\
	 3 & Take a taxi to Kucuk Ayasofya Camii (6 min), stay for 0.14 hour  \\
	 4 & Walk to Blue Mosque (6 min), stay for 0.90 hour  \\
	 5 & Walk to Basilica Cistern (4 min), stay for 0.90 hour  \\
	 6 & Walk to Hagia Sophia Museum (4 min), stay for 0.93 hour  \\
	 7 & Walk to Gulhane Park (4 min), stay for 0.11 hour  \\
	 8 & Walk to Archaeological Museums (2 min), stay for 0.78 hour  \\
	 9 & Take a taxi to Rustem Pasha Mosque (6 min), stay for 0.78 hour  \\
	 10 & Take a taxi to Rahmi M. Koc Museum (9 min), stay for 0.76 hour  \\
	 11 & Take a taxi to Kariye Museum (8 min), stay for 0.82 hour  \\
	 12 & Take a taxi and return to Suleymaniye Mosque (12 min) 	 \vspace*{2mm}
\\
	 \hline\hline
	 \end{tabularx}
\end{center}
\end{table}

During a recent trip to Istanbul for the WAFR 2014 conference, due to a tight schedule, some of us only had a few hours to visit local attractions. In the end, we visited the Hagia Sophia Museum, the Blue Mosque, and the Basilica Cistern. It turns out that, when we run the \rmt\, algorithm with three hours of budget, this is the exact itinerary returned by the algorithm (see Table~\ref{table:itinerary-3}). 

\begin{table}[htp]
\begin{center}
\caption{\label{table:itinerary-3} A 3-hour computed itinerary in Istanbul. }
	\begin{tabularx}{\columnwidth}{cX}
	 \hline\hline	 \\
	 1 & Start at Basilica Cistern, stay for 0.90 hour  \\
	 2 & Walk to Hagia Sophia Museum (4 min), stay for 0.93 hour  \\
	 3 & Walk to Blue Mosque (8 min), stay for 0.89 hour  \\
	 4 & Walk back to Basilica Cistern (4 min) 	 \vspace*{2mm}
\\
	 \hline\hline
	 \end{tabularx}
\end{center}
\end{table}

\section{Conclusion}\label{section:conclusion}
In this paper, we proposed the Optimal Tourist Problem (\otp) that tie together the problem of maximizing information collection efforts at point-of-interests (POIs) and minimizing the required time spent on traveling between the set of discrete, distributed POIs. A particular novelty is that our formulation encompasses a general class of time-based reward functions. For solving the two variants of \otp, \rmt\, and \bmt, we construct an exact (when reward function is linear) or an arbitrarily optimal (when reward function is non-linear) MIP model that gives rise to an anytime algorithm for solving such problems. Computational results suggest that our algorithm is applicable to practical-sized itinerary planning or informative path planning problems and generates fairly sensible plans.

\bibliographystyle{IEEEtranN}
\bibliography{../../../../references/jingjin}

\begin{thebibliography}{24}
\providecommand{\natexlab}[1]{#1}
\providecommand{\url}[1]{#1}
\csname url@samestyle\endcsname
\providecommand{\newblock}{\relax}
\providecommand{\bibinfo}[2]{#2}
\providecommand{\BIBentrySTDinterwordspacing}{\spaceskip=0pt\relax}
\providecommand{\BIBentryALTinterwordstretchfactor}{4}
\providecommand{\BIBentryALTinterwordspacing}{\spaceskip=\fontdimen2\font plus
\BIBentryALTinterwordstretchfactor\fontdimen3\font minus
  \fontdimen4\font\relax}
\providecommand{\BIBforeignlanguage}[2]{{%
\expandafter\ifx\csname l@#1\endcsname\relax
\typeout{** WARNING: IEEEtranN.bst: No hyphenation pattern has been}%
\typeout{** loaded for the language `#1'. Using the pattern for}%
\typeout{** the default language instead.}%
\else
\language=\csname l@#1\endcsname
\fi
#2}}
\providecommand{\BIBdecl}{\relax}
\BIBdecl

\bibitem[Smith et~al.(2011)Smith, Schwager, Smith, Jones, Rus, and
  Sukhatme]{SmiSchSmiJonRusSuk11}
R.~N. Smith, M.~Schwager, S.~L. Smith, B.~H. Jones, D.~Rus, and G.~S. Sukhatme,
  ``Persistent ocean monitoring with underwater gliders: Adapting sampling
  resolution,'' \emph{Journal of Field Robotics}, vol.~28, no.~5, pp. 714--741,
  Sep-Oct 2011.

\bibitem[Grocholsky et~al.(2006)Grocholsky, Keller, Kumar, and
  Pappas]{GroKelKumPap06}
B.~Grocholsky, J.~Keller, V.~Kumar, and G.~Pappas, ``Cooperative air and ground
  surveillance,'' \emph{IEEE Robotics and Automation Magazine}, vol.~13, no.~3,
  pp. 16--25, Sep 2006.

\bibitem[Alamdari et~al.(2014)Alamdari, Fata, and Smith]{AlaFatSmi14}
S.~Alamdari, E.~Fata, and S.~L. Smith, ``Persistent monitoring in discrete
  environments: Minimizing the maximum weighted latency between observations,''
  \emph{The International Journal of Robotics Research}, vol.~33, no.~1, pp.
  138--154, 2014.

\bibitem[Smith et~al.(2012)Smith, Schwager, and Rus]{SmiSchRus12}
S.~L. Smith, M.~Schwager, and D.~Rus, ``Persistent robotic tasks: Monitoring
  and sweeping in changing environments,'' \emph{IEEE Transactions on
  Robotics}, vol.~28, no.~2, pp. 410--426, April 2012.

\bibitem[Yu et~al.(2014)Yu, Karaman, and Rus]{YuKarRus14ICRA}
J.~Yu, S.~Karaman, and D.~Rus, ``Persistent monitoring of events with
  stochastic arrivals at multiple stations,'' in \emph{Proceedings IEEE
  International Conference on Robotics \& Automation}, 2014, preliminary
  extended journal version available at http://arxiv.org/abs/1309.6041.

\bibitem[Lim et~al.(2014)Lim, Hsu, and Lee]{LimHsuLee14}
Z.~W. Lim, D.~Hsu, and W.~S. Lee, ``Adaptive informative path planning in
  metric spaces,'' in \emph{Proceedings Workshop on Algorithmic Foundations of
  Robotics}, 2014.

\bibitem[Kavraki et~al.(1996)Kavraki, Svestka, Latombe, and
  Overmars]{KavSveLatOve96}
L.~E. Kavraki, P.~Svestka, J.-C. Latombe, and M.~H. Overmars, ``Probabilistic
  roadmaps for path planning in high-dimensional configuration spaces,''
  \emph{IEEE Transactions on Robotics \& Automation}, vol.~12, no.~4, pp.
  566--580, Jun. 1996.

\bibitem[LaValle(1998)]{Lav98c}
S.~M. LaValle, ``Rapidly-exploring random trees: {A} new tool for path
  planning,'' Iowa State University, Tech. Rep., Oct 1998, computer Science
  Department TR 98-11.

\bibitem[Karaman and Frazzoli(2011)]{KarFra11IJRR}
\BIBentryALTinterwordspacing
S.~Karaman and E.~Frazzoli, ``Sampling-based algorithms for optimal motion
  planning,'' \emph{International Journal of Robotics Research}, vol.~30,
  no.~7, pp. 846--894, June 2011. [Online]. Available:
  \url{http://ares.lids.mit.edu/papers/Karaman.Frazzoli.IJRR11.pdf}
\BIBentrySTDinterwordspacing

\bibitem[Hollinger and Sukhatme(2013)]{HolSuk13}
G.~A. Hollinger and G.~S. Sukhatme, ``Sampling-based motion planning for
  robotic information gathering,'' in \emph{Robotics: Science and Systems},
  2013.

\bibitem[Lan and Schwager(2013)]{LanSch13}
X.~Lan and M.~Schwager, ``Planning periodic persistent monitoring trajectories
  for sensing robots in gaussian random fields,'' in \emph{Proceedings IEEE
  International Conference on Robotics \& Automation}, May 2013, pp.
  2407--2412.

\bibitem[Chao et~al.(1996)Chao, Golden, and Wasil]{ChaGolWas96a}
I.~Chao, B.~Golden, and E.~Wasil, ``Theory and methodology - the team
  orienteering problem,'' \emph{European Journal of Operational Research},
  vol.~88, pp. 464--474, 1996.

\bibitem[Vansteenwegen et~al.(2011)Vansteenwegen, Souffriau, and
  Oudheusden]{VanSouVan11}
P.~Vansteenwegen, W.~Souffriau, and D.~V. Oudheusden, ``The orienteering
  problem: A survey,'' \emph{European Journal of Operational Research}, vol.
  209, pp. 1--10, 2011.

\bibitem[Gavalas et~al.(2014)Gavalas, Konstantopoulos, Mastakas, and
  Pantziou]{GavKonMasPan14}
D.~Gavalas, C.~Konstantopoulos, J.~Mastakas, and G.~Pantziou, ``A survey on
  algorithmic approaches for solving tourist trip design problems,''
  \emph{Journal of Heuristics}, vol.~20, no.~3, pp. 291--328, 2014.

\bibitem[Chekuri et~al.(2012)Chekuri, Korula, and P{\'a}l]{CheKorPal12}
C.~Chekuri, N.~Korula, and M.~P{\'a}l, ``Improved algorithms for orienteering
  and related problems,'' \emph{ACM Transactions on Algorithms (TALG)}, vol.~8,
  no.~3, p.~23, 2012.

\bibitem[Erdo\v{g}an and Laporte(2013)]{ErdLap13}
G.~Erdo\v{g}an and G.~Laporte, ``The orienteering problem with variable
  profits,'' \emph{Networks}, vol.~61, no.~2, pp. 104--116, 2013.

\bibitem[De~Choudhury et~al.(2010)De~Choudhury, Feldman, Amer-Yahia, Golbandi,
  Lempel, and Yu]{ChoFelAme10}
M.~De~Choudhury, M.~Feldman, S.~Amer-Yahia, N.~Golbandi, R.~Lempel, and C.~Yu,
  ``Automatic construction of travel itineraries using social breadcrumbs,'' in
  \emph{Proceedings of the 21st ACM conference on Hypertext and
  hypermedia}.\hskip 1em plus 0.5em minus 0.4em\relax ACM, 2010, pp. 35--44.

\bibitem[Basu~Roy et~al.(2011)Basu~Roy, Das, Amer-Yahia, and Yu]{BasDasAmeYu11}
S.~Basu~Roy, G.~Das, S.~Amer-Yahia, and C.~Yu, ``Interactive itinerary
  planning,'' in \emph{Data Engineering (ICDE), 2011 IEEE 27th International
  Conference on}.\hskip 1em plus 0.5em minus 0.4em\relax IEEE, 2011, pp.
  15--26.

\bibitem[Yoon et~al.(2012)Yoon, Zheng, Xie, and Woo]{HyoZheXieWoo12}
H.~Yoon, Y.~Zheng, X.~Xie, and W.~Woo, ``Social itinerary recommendation from
  user-generated digital trails,'' \emph{Personal and Ubiquitous Computing},
  vol.~16, no.~5, pp. 469--484, 2012.

\bibitem[Chekuri and P{\'a}l(2005)]{ChePal05}
C.~Chekuri and M.~P{\'a}l, ``A recursive greedy algorithm for walks in directed
  graphs,'' in \emph{Foundations of Computer Science, 2005. FOCS 2005. 46th
  Annual IEEE Symposium on}.\hskip 1em plus 0.5em minus 0.4em\relax IEEE, 2005,
  pp. 245--253.

\bibitem[Floyd(1962)]{Flo62}
R.~W. Floyd, ``Algorithm 97: shortest path,'' \emph{Communications of the ACM},
  vol.~5, no.~6, p. 345, 1962.

\bibitem[Warshall(1962)]{War62}
S.~Warshall, ``A theorem on boolean matrices,'' \emph{Journal of the ACM
  (JACM)}, vol.~9, no.~1, pp. 11--12, 1962.

\bibitem[Land and Doig(1960)]{LanDoi60}
A.~H. Land and A.~G. Doig, ``An automatic method of solving discrete
  programming problems,'' \emph{Econometrica}, vol.~28, no.~3, pp. 497--520,
  1960.

\bibitem[Gurobi~Optimization(2014)]{gurobi}
\BIBentryALTinterwordspacing
I.~Gurobi~Optimization, ``Gurobi optimizer reference manual,'' 2014. [Online].
  Available: \url{http://www.gurobi.com}
\BIBentrySTDinterwordspacing

\end{thebibliography}

\end{document}